\documentclass[10pt,twocolumn,letterpaper]{article}

\usepackage{caption}  % import this before manually setting captions in wacv
\usepackage{wacv}
\usepackage{times}
\usepackage{epsfig}
\usepackage{graphicx}
\usepackage{amsmath}
\usepackage{amssymb}

\makeatletter
\@namedef{ver@everyshi.sty}{}
\makeatother
\usepackage{tikz}
\usepackage{cite}
\usepackage{booktabs}
\usepackage{multirow}
\newcommand{\ra}[1]{\renewcommand{\arraystretch}{#1}}
\usepackage{pgfplots}
\usepackage[export]{adjustbox}
\pgfplotsset{compat=1.17}
\usepackage{siunitx}
\DeclareSIUnit[number-unit-product = ]\k{k}
\usepackage[mode=buildnew]{standalone}
\usepackage{calc}
\usepackage{xr-hyper}
\usepackage{algorithm}
\usepackage{algcompatible}
\usetikzlibrary{calc}
\usepackage{filecontents}
\usepackage{colortbl}
\usepackage{pgfplotstable}
\usetikzlibrary{pgfplots.colorbrewer}

\DeclareMathOperator*{\argmax}{argmax}

\DeclareMathOperator*{\mean}{mean}

\newcommand{\textover}[3][l]{%
 \makebox[\widthof{#3}][#1]{#2}%
}

\newcommand{\PAR}[1]{\vskip2pt \noindent{\bf #1}}

\colorlet{shadecolor}{gray!20}

\makeatletter
\tikzset{%
  remember picture with id/.style={%
    remember picture,
    overlay,
    save picture id=#1,
  },
  save picture id/.code={%
    \edef\pgf@temp{#1}%
    \immediate\write\pgfutil@auxout{%
      \noexpand\savepointas{\pgf@temp}{\pgfpictureid}}%
  },
  if picture id/.code args={#1#2#3}{%
    \@ifundefined{save@pt@#1}{%
      \pgfkeysalso{#3}%
    }{
      \pgfkeysalso{#2}%
    }
  }
}

\def\savepointas#1#2{%
  \expandafter\gdef\csname save@pt@#1\endcsname{#2}%
}

\def\tmk@labeldef#1,#2\@nil{%
  \def\tmk@label{#1}%
  \def\tmk@def{#2}%
}

\tikzdeclarecoordinatesystem{pic}{%
  \pgfutil@in@,{#1}%
  \ifpgfutil@in@%
    \tmk@labeldef#1\@nil
  \else
    \tmk@labeldef#1,(0pt,0pt)\@nil
  \fi
  \@ifundefined{save@pt@\tmk@label}{%
    \tikz@scan@one@point\pgfutil@firstofone\tmk@def
  }{%
  \pgfsys@getposition{\csname save@pt@\tmk@label\endcsname}\save@orig@pic%
  \pgfsys@getposition{\pgfpictureid}\save@this@pic%
  \pgf@process{\pgfpointorigin\save@this@pic}%
  \pgf@xa=\pgf@x
  \pgf@ya=\pgf@y
  \pgf@process{\pgfpointorigin\save@orig@pic}%
  \advance\pgf@x by -\pgf@xa
  \advance\pgf@y by -\pgf@ya
  }%
}
\newlength\AlgIndent
\newcounter{mymark}
\newcommand\ColorLine{%
  \stepcounter{mymark}%
  \tikz[remember picture with id=mark-\themymark,overlay] {;}%
  \begin{tikzpicture}[remember picture,overlay]%
    \filldraw[shadecolor]%
   let \p1=(pic cs:mark-\themymark), 
   \p2=(current page.east)  in 
   ([xshift=-\ALG@thistlm,yshift=-1.0ex]0,\y1)  rectangle ++(\linewidth+\AlgIndent,\baselineskip+0.2ex);
  \end{tikzpicture}%
}%

\algnewcommand\CREQUIRE{\item[\setlength\AlgIndent{1.6em}\ColorLine\algorithmicrequire]}%
\algnewcommand\CENSURE{\item[\setlength\AlgIndent{1.6em}\ColorLine\algorithmicensure]}%
\algnewcommand\CSTATE{\State\ColorLine}%
\algnewcommand\CSTATEx{\Statex\ColorLine}%
\algnewcommand\CCOMMENT{\Comment\ColorLine}%

\algdef{SE}[WHILE]{CWHILE}{ENDWHILE}%
   [2][default]{\ColorLine\algorithmicwhile\ #2\ \algorithmicdo\ALG@compatcomm{#1}}%
   {\algorithmicend\ \algorithmicwhile}%
\algdef{SE}[FOR]{CFOR}{ENDFOR}%
   [2][default]{\ColorLine\algorithmicfor\ #2\ \algorithmicdo\ALG@compatcomm{#1}}%
   {\algorithmicend\ \algorithmicfor}%
\algdef{S}[FOR]{CFORALL}%
   [2][default]{\ColorLine\algorithmicforall\ #2\ \algorithmicdo\ALG@compatcomm{#1}}%
\algdef{SE}[LOOP]{CLOOP}{ENDLOOP}%
   [1][default]{\ColorLine\algorithmicloop\ALG@compatcomm{#1}}%
   {\algorithmicend\ \algorithmicloop}%
\algdef{SE}[REPEAT]{CREPEAT}{UNTIL}%
   [1][default]{\ColorLine\algorithmicrepeat\ALG@compatcomm{#1}}%
   [1]{\algorithmicuntil\ #1}%
\algdef{SE}[IF]{CIF}{ENDIF}%
   [2][default]{\ColorLine\algorithmicif\ #2\ \algorithmicthen\ALG@compatcomm{#1}}%
   {\ColorLine\algorithmicend\ \algorithmicif}%
\algdef{C}[IF]{IF}{CELSIF}%
   [2][default]{\ColorLine\algorithmicelse\ \algorithmicif\ #2\ \algorithmicthen\ALG@compatcomm{#1}}%
\algdef{Ce}[ELSE]{IF}{CELSE}{ENDIF}%
   [1][default]{\ColorLine\algorithmicelse\ALG@compatcomm{#1}}%
    
\makeatother

\algrenewcommand\alglinenumber[1]{\tiny #1:}
\newcommand{\algoname}[1]{\textnormal{\textsc{#1}}}
\renewcommand{\algorithmiccomment}[1]{\bgroup\hfill\footnotesize//~#1\egroup}

\usepackage[page]{appendix}

\newcommand{\beginappendixa}{%
        \setcounter{table}{0}
        \renewcommand{\thetable}{A-\arabic{table}}%
        \setcounter{figure}{0}
        \renewcommand{\thefigure}{A-\arabic{figure}}%
     }
\newcommand{\beginappendixb}{%
        \setcounter{table}{0}
        \renewcommand{\thetable}{B-\arabic{table}}%
        \setcounter{figure}{0}
        \renewcommand{\thefigure}{B-\arabic{figure}}%
     }
\newcommand{\beginappendixc}{%
        \setcounter{table}{0}
        \renewcommand{\thetable}{C-\arabic{table}}%
        \setcounter{figure}{0}
        \renewcommand{\thefigure}{C-\arabic{figure}}%
     }
\newcommand{\beginappendixd}{%
        \setcounter{table}{0}
        \renewcommand{\thetable}{D-\arabic{table}}%
        \setcounter{figure}{0}
        \renewcommand{\thefigure}{D-\arabic{figure}}%
     }
\newcommand{\beginappendixe}{%
        \setcounter{table}{0}
        \renewcommand{\thetable}{E-\arabic{table}}%
        \setcounter{figure}{0}
        \renewcommand{\thefigure}{E-\arabic{figure}}%
     }
\newcommand{\beginappendixf}{%
        \setcounter{table}{0}
        \renewcommand{\thetable}{F-\arabic{table}}%
        \setcounter{figure}{0}
        \renewcommand{\thefigure}{F-\arabic{figure}}%
     }

\def\wacvPaperID{21} % Enter the WACV Paper ID here

\wacvalgorithmstrack   % Uncomment this line if you are submitting to the Algorithms Track.
\wacvfinalcopy % *** Uncomment this line for the final submission

\ifwacvfinal
\usepackage[hidelinks,breaklinks=true,bookmarks=false]{hyperref}
\else
\usepackage[pagebackref=true,breaklinks=true,colorlinks,bookmarks=false]{hyperref}
\fi

\begin{document}

%%%%%%%%% TITLE
\title{Refign: Align and Refine for Adaptation of Semantic Segmentation\\to Adverse Conditions}

\author{David Bruggemann
\qquad Christos Sakaridis
\qquad Prune Truong
\qquad Luc Van Gool \\
ETH Zurich, Switzerland \\
{\tt\small \{brdavid, csakarid, truongp, vangool\}@vision.ee.ethz.ch}
% For a paper whose authors are all at the same institution,
% omit the following lines up until the closing ``}''.
% Additional authors and addresses can be added with ``\and'',
% just like the second author.
% To save space, use either the email address or home page, not both
}

\maketitle
\thispagestyle{empty}

%%%%%%%%% ABSTRACT
\begin{abstract}
Due to the scarcity of dense pixel-level semantic annotations for images recorded in adverse visual conditions, there has been a keen interest in unsupervised domain adaptation (UDA) for the semantic segmentation of such images. UDA adapts models trained on normal conditions to the target adverse-condition domains.
Meanwhile, multiple datasets with driving scenes provide corresponding images of the same scenes across multiple conditions, which can serve as a form of weak supervision for domain adaptation.
We propose Refign, a generic extension to self-training-based UDA methods which leverages these cross-domain correspondences.
Refign consists of two steps:
(1) aligning the normal-condition image to the corresponding adverse-condition image using an uncertainty-aware dense matching network, and
(2) refining the adverse prediction with the normal prediction using an adaptive label correction mechanism.
We design custom modules to streamline both steps and set the new state of the art for domain-adaptive semantic segmentation on several adverse-condition benchmarks, including ACDC and Dark Zurich.
The approach introduces no extra training parameters, minimal computational overhead---during training only---and can be used as a drop-in extension to improve any given self-training-based UDA method.
Code is available at \url{https://github.com/brdav/refign}.
\end{abstract}

%%%%%%%%% INTRODUCTION
\section{Introduction}

\begin{figure}
\centering
    \includegraphics[width=\linewidth]{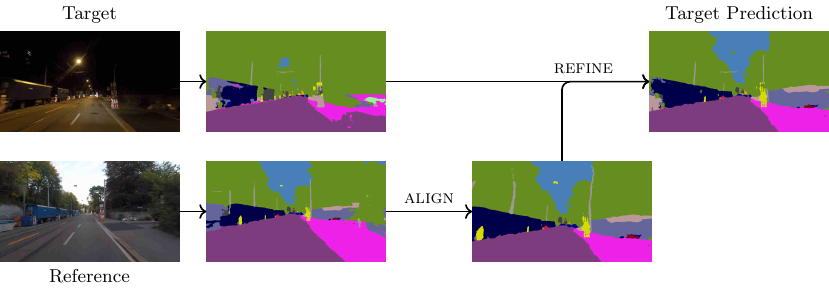}
    \caption{By leveraging a reference image depicting a similar scene as the target, Refign improves target predictions in two steps: (1) The reference predictions are spatially aligned with the target. (2) The target predictions are refined via adaptive label correction.}
    \label{fig:teaser}
    \vspace{-3mm}
\end{figure}

Semantic segmentation is a central task for scene understanding, \eg, in fully autonomous vehicle systems.
In such safety-critical applications, robustness of the segmentation model to adverse visual conditions is pivotal.
Since state-of-the-art semantic segmentation models are typically trained on clear-weather domains~\cite{cordts2016cityscapes}, where detailed pixel-level annotations are available, they have proven to be frail~\cite{kamann2020benchmarking,sakaridis2021acdc} with respect to changes in image quality, illumination, or weather.
Accordingly, a large body of research has focused on unsupervised domain adaptation (UDA) to adapt these models to different domains in which no labels are available~\cite{FCNs:in:the:wild,tsai2018learning,vu2019advent,yang2020fda,SynRealDataFogECCV18}.

In this paper, we propose an extension to UDA approaches, which leverages additional \emph{reference}\textemdash or normal-condition\textemdash images to improve the predictions of the target-domain images (see Fig.~\ref{fig:teaser}).
The reference image depicts the same scene as the target image, albeit from a different viewpoint and under favorable conditions (daytime and clear weather).
For driving datasets, such image pairs can be collected with minimal extra effort by capturing the same route twice and matching frames via GPS.
In recent years, a number of driving datasets have followed this procedure, \eg RobotCar~\cite{maddern20171}, Dark Zurich~\cite{sakaridis2020map}, ACDC~\cite{sakaridis2021acdc}, and Boreas~\cite{burnett2022boreas}.
When adapting a semantic segmentation model from a normal-condition dataset (\eg Cityscapes~\cite{cordts2016cityscapes}) to adverse conditions, the reference frames represent an \emph{intermediate} domain.
While they overlap with the target frames in terms of sensor and region characteristics, they share source domain weather and time of day.
This can bolster the domain adaptation process by providing complementary, more easily learnable information, even if reference and target images might differ slightly in semantic content.

Current state-of-the-art UDA approaches~\cite{zhang2021prototypical,hoyer2021daformer,xie2022sepico} rely on self-training~\cite{yarowsky1995unsupervised}, where the network is trained with its own target domain predictions as self-supervision.
The surprising effectiveness of self-training can be largely attributed to clever regularization strategies~\cite{wei2020theoretical}, in the absence of which it suffers from confirmation bias.
These regularization strategies aim at correctly propagating available true labels to neighboring unlabeled samples in an iterative fashion.
A critical issue in this procedure is the error propagation of noisy labels, leading to a drift in pseudo-labels if unmitigated.
It has been shown that large neural networks easily overfit to label noise, which deteriorates their generalization performance~\cite{arpit2017closer,zhang2021understanding}.
Our method ameliorates this error propagation issue by incorporating the predictions of two separate \emph{views} to reason about the labels of a given scene.
Broadly speaking, it could thus be considered an instance of Multi-View Learning~\cite{yan2021deep}.
More specifically, we consider the target prediction as a noisy label to be modified by the complementary reference class-wise probabilities, posing the fusion process as self-label correction~\cite{wang2021proselflc}.
Echoing recent advances in that field, we design an adaptive label refinement mechanism, which allows the learner to ignore or revise noisy labels.

Considering that semantic segmentation requires precise, pixel-level predictions, we hypothesize that the reference-target label fusion benefits greatly from spatial alignment of the two frames.
Thus, in a preliminary step to refinement, we warp the reference prediction to align it with the target.
In anticipation that such alignment is imperfect\textemdash due to dynamic objects, occlusions, and warp inaccuracies\textemdash we jointly estimate a confidence for each warped pixel, which serves as guidance in the downstream refinement.
To streamline this process, we design a probabilistic extension to the geometric matching framework WarpC~\cite{truong2021warp}, and show its effectiveness in terms of accuracy and uncertainty awareness.

Altogether, Refign consists of the alignment module and the refinement module.
Both modules introduce limited computational overhead during training and are shown to boost the performance of baseline UDA methods significantly.
When added on top of DAFormer~\cite{hoyer2021daformer}, Refign achieves 65.6\% and 56.2\%\,mIoU for semantic segmentation on ACDC and Dark Zurich respectively, setting the new state of the art on these adverse-condition benchmarks.

%%%%%%%%% RELATED WORK
\section{Related Work}

\PAR{Adaptation to Adverse Domains.}
Several works on domain adaptation for semantic segmentation have been developed with the synthetic-to-real adaptation setting in mind, focusing on adversarial alignment of the source and target domains~\cite{FCNs:in:the:wild,cyCADA,tsai2018learning,synthetic:semantic:segmentation,FCNs:adaptation,vu2019advent,luo2019taking,wang2020differential,patch:align:adaptation}, self-training with pseudo-labels in the target domain~\cite{self:training:adaptation,zou2019confidence}, and combining self-training with adversarial adaptation~\cite{li2019bidirectional} or with pixel-level adaptation via explicit transforms from source to target~\cite{yang2020fda,textda:adaptation}. Significantly fewer methods have been presented for adaptation from \emph{normal to adverse} domains, which is highly relevant for practical scenarios such as automated driving, in which the perception system needs to be robust to unfavorable conditions such as fog, night, rain, and snow. Generating partially synthetic data with simulated adverse weather from clear-weather counterparts has proven to improve performance on real adverse-condition sets featuring fog~\cite{sakaridis2018semantic} and rain~\cite{tremblay2021rain}. Sequences of domains of progressively increasing adversity have been leveraged in~\cite{SynRealDataFogECCV18,dai2020curriculum,incremental:adversarial:DA:18} via curriculum-based schemes. Light-weight input adapters~\cite{porav2019don} and adversarial style transfer~\cite{romera2019bridging,sun2019see} are proposed as a generic pre-processing step at test time before predicting with source-domain models.
Shared characteristics between different domains, such as sensor and time of day~\cite{gao2022cross} or visibility~\cite{lee2022fifo}, are leveraged to learn consistent representations across different datasets.
The recent introduction of adverse-condition semantic segmentation datasets with image-level cross-condition correspondences, such as Dark Zurich~\cite{sakaridis2019guided} and ACDC~\cite{sakaridis2021acdc}, has enabled the development of approaches which leverage these correspondences as weak supervision for adapting to adverse conditions.
Sparse, pixel-level correspondences are used in~\cite{larsson2019cross} to enforce consistency of the predictions across different conditions.
DANIA~\cite{wu2021one} warps daytime predictions into the nighttime image viewpoint and applies a consistency loss for static classes only.
Closely related to our work, MGCDA~\cite{sakaridis2020map} \emph{fuses} cross-time-of-day predictions after two-view-geometry-based alignment.
Differently from their work, we directly warp the two corresponding images with an uncertainty-aware dense matching network.
The warp uncertainty provides guidance for the downstream fusion, which enables more nuanced refinement, even incorporating dynamic objects.
Finally, while most of the aforementioned approaches are tailored to specific conditions, our method can address arbitrary adverse conditions.

\PAR{Dense Geometric Matching.}
Dense correspondence estimation aims at finding pixel-wise matches relating a pair of images. Approaches such as~\cite{melekhov2019dgc, rocco2018neighbourhood,li2020dual,truong2022probabilistic} predict a 4D correlation volume, from which the dense correspondences are extracted as the $\argmax$ of the matching scores. 
Our matching network instead follows a recent line of works~\cite{melekhov2019dgc,jiang2021cotr,truong2021pdc,truong2020glu,truong2021warp,shen2020ransac} which directly regress the dense flow field or correspondence map. DGC-Net~\cite{melekhov2019dgc} employs a coarse-to-fine approach where a global cost volume is constructed at the coarsest scale to handle large motions. However, it can only handle input images of a fixed, low resolution, which significantly impacts the accuracy of the predicted flow. To circumvent this issue, GLU-Net~\cite{truong2020glu} integrates both local and global correlation layers. RANSAC-Flow~\cite{shen2020ransac} is based on a two-stage refinement strategy, first estimating homographies relating the pair and then refining them by a predicted residual flow. COTR~\cite{jiang2021cotr} relies on a transformer-based architecture to perform matching. Similarly to us, PDC-Net~\cite{truong2021pdc} adopts a probabilistic framework to regress both the flow and its uncertainty.
Differently from our work, it requires sparse ground-truth matches obtained through structure from motion (SfM) for training.
Instead, we present a probabilistic extension of the warp consistency framework~\cite{truong2021warp}, which leverages both synthetic flow and real image pairs and requires no training annotations.

\PAR{Label Correction.}
Label correction (LC) approaches aim to improve learning from noisy labels by modifying the one-hot labels, \eg through convex combinations with predicted distributions.
Response-based knowledge distillation (KD)~\cite{gou2021knowledge} methods are a prominent example for LC.
While the correction is initiated by a separate teacher model in KD, Self-LC methods~\cite{reed2014training,tanaka2018joint,song2019selfie,yuan2020revisiting,wang2021proselflc} rely on the learner itself to correct erroneous labels.
Similarly to our method, multiple works have used Self-LC to improve self-training for domain-adaptive semantic segmentation~\cite{zheng2019unsupervised,zheng2021rectifying,zhang2021prototypical}.
In contrast to these works, our method uses complementary information gained from two different \emph{views} of the same scene to correct the labels.

%%%%%%%%% METHOD
\section{Refign}
\label{sec:method}

\begin{figure}
\centering
\includegraphics[width=\linewidth]{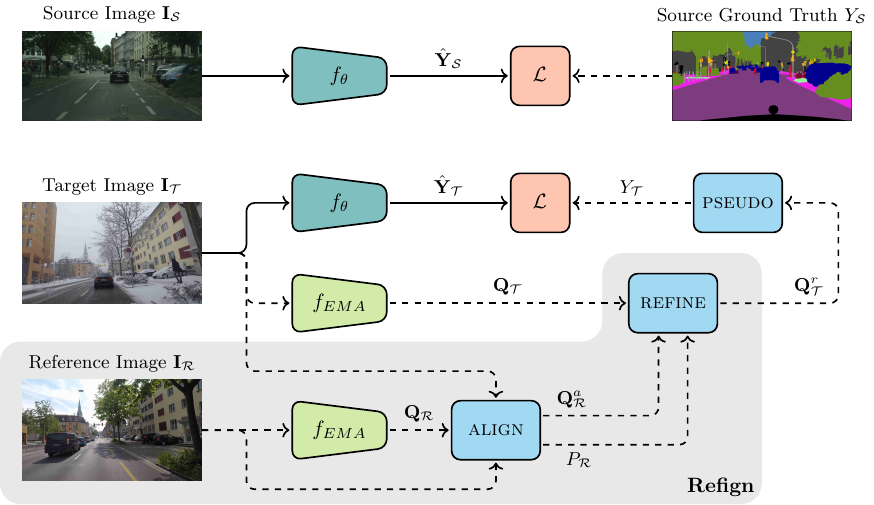}
    \caption{Generic self-training for UDA, complemented by the proposed Refign (gray-shaded area). Refign leverages an additional reference image to improve target pseudo-labels with two modules. The \textnormal{\textsc{align}} module is a pre-trained dense matching network (Sec.~\ref{sec:method:alignment}) which (1) warps the reference prediction to align it with the target ($\mathbf{Q}^a_\mathcal{R}$) and (2) estimates a confidence map of the warp ($P_\mathcal{R}$). The \textnormal{\textsc{refine}} module (Sec.~\ref{sec:method:refinement}) combines the above quantities to improve the target pseudo-labels $Y_\mathcal{T}$ for self-training. 
    Only solid arrows backpropagate gradients.}
    \label{fig:method_overview}
    \vspace{-5mm}
\end{figure}

\begin{algorithm}[t]
\footnotesize
\setlength{\fboxsep}{1pt}
\caption{Generic UDA self-training, complemented with Refign (shaded in gray).} 
\label{alg:refign}

\begin{algorithmic}[1]
\REQUIRE Samples $\mathcal{D}_{\mathcal{S}}$, $\mathcal{D}_{\mathcal{T}}$, $\mathcal{D}_{\mathcal{R}}$, initialized network $f_\theta$, pretrained alignment module \algoname{align}
\FOR{$i=0$ to $N$} 
  \State update/initialize teacher network $f_{EMA}$ %\Comment{Update weights of $f_{EMA}$}
  \State sample $(\mathbf{I}_\mathcal{S}, Y_\mathcal{S})$ from $\mathcal{D}_{\mathcal{S}}$, $\mathbf{I}_\mathcal{T}$ from $\mathcal{D}_{\mathcal{T}}$\colorbox{shadecolor}{\strut, and $\mathbf{I}_\mathcal{R}$ from $\mathcal{D}_{\mathcal{R}}$}
  \State $\hat{\mathbf{Y}}_\mathcal{S} \leftarrow f_\theta(\mathbf{I}_\mathcal{S})$
  \CIF{$\algoname{rand}(0,1) < 0.5$} \Comment{Adapt to $\mathcal{T}$}
    \State $\mathbf{Q}_\mathcal{T} \leftarrow f_{EMA}(\mathbf{I}_\mathcal{T})$ \colorbox{shadecolor}{and $\mathbf{Q}_\mathcal{R}\leftarrow f_{EMA}(\mathbf{I}_\mathcal{R})$}
    \CSTATE $\mathbf{Q}^a_\mathcal{R},P_\mathcal{R} \leftarrow \algoname{align}(\mathbf{Q}_\mathcal{R},\mathbf{I}_\mathcal{R},\mathbf{I}_\mathcal{T})$  \Comment{Warp $\mathbf{Q}_\mathcal{R}$} \label{alglst:align}
    \CSTATE $\mathbf{Q}^r_\mathcal{T} \leftarrow \algoname{refine}(\mathbf{Q}_\mathcal{T},\mathbf{Q}^a_\mathcal{R},P_\mathcal{R})$  \Comment{Eq.~\ref{eq:convex_comb}} \label{alglst:refine}
    \State $Y_\mathcal{T} \leftarrow \algoname{pseudo}(\mathbf{Q}^r_\mathcal{T})$ \Comment{\eg with $\argmax$} \label{alglst:pseudo}
    \State $\hat{\mathbf{Y}}_\mathcal{T} \leftarrow f_\theta(\mathbf{I}_\mathcal{T})$
    \State $l = \mathcal{L}(\hat{\mathbf{Y}}_\mathcal{S},Y_\mathcal{S}) + \mathcal{L}(\hat{\mathbf{Y}}_\mathcal{T},Y_\mathcal{T})$
  \CELSE \Comment{Adapt to $\mathcal{R}$}  \label{alglst:adr_start}
    \CSTATE $Y_\mathcal{R} \leftarrow \algoname{pseudo}(f_{EMA}(\mathbf{I}_\mathcal{R}))$
    \CSTATE $\hat{\mathbf{Y}}_\mathcal{R} \leftarrow f_\theta(\mathbf{I}_\mathcal{R})$
    \CSTATE $l = \mathcal{L}(\hat{\mathbf{Y}}_\mathcal{S},Y_\mathcal{S}) + \mathcal{L}(\hat{\mathbf{Y}}_\mathcal{R},Y_\mathcal{R})$  \label{alglst:adr_end}
  \ENDIF
  \State apply gradient descent to $\theta$ with $\nabla_\theta l$
\ENDFOR
\end{algorithmic}

% \end{document}
\end{algorithm}

Given labeled images from a source domain $\mathcal{S}$ (\eg Cityscapes~\cite{cordts2016cityscapes}), and unlabeled, corresponding pairs of images from a target domain $\mathcal{T}$ (\eg adverse-condition images of ACDC~\cite{sakaridis2021acdc}) \emph{and} a reference domain $\mathcal{R}$ (\eg normal-condition images of ACDC), we aim to learn a model which predicts semantic segmentation maps for target-domain images. 
Ground-truth semantic segmentation labels are only available for the source-domain images during training.
The reference image is assumed to depict the same scene as the target image, but from a different viewpoint and under better visual conditions.

Our method\textemdash Refign\textemdash is a framework-agnostic extension to self-training-based UDA methods, leveraging the additional reference image.
Underpinning Refign is the hypothesis that $\mathcal{R}$ can serve as an intermediate domain between $\mathcal{S}$ and $\mathcal{T}$.
It is a well-established fact that intermediate domains can bolster UDA~\cite{gopalan2011domain,gong2012geodesic,SynRealDataFogECCV18}.
In our case, we hypothesize that the higher-quality predictions for $\mathcal{R}$ can be used to guide self-training in $\mathcal{T}$.

Fig.~\ref{fig:method_overview} shows a generic self-training UDA setup, with our Refign module shaded in gray. 
In each training iteration, the model $f_\theta$ is trained both with the source ground-truth labels $Y_\mathcal{S}$ and the target pseudo-labels $Y_\mathcal{T}$.
Most state-of-the-art UDA approaches~\cite{hoyer2021daformer,xie2022sepico,hoyer2022hrda} generate the pseudo-labels with a Mean Teacher~\cite{tarvainen2017mean} ($f_{EMA}$), using the exponential moving average (EMA) weights of $f_\theta$.
This increases pseudo-label accuracy and mitigates confirmation bias~\cite{tarvainen2017mean}.
As depicted in Fig.~\ref{fig:method_overview} and summarized in Alg.~\ref{alg:refign}, Refign introduces two additional steps at training time to improve the pseudo-labels:
(1) A pre-trained alignment module (Line~\ref{alglst:align}) computes the flow from target to reference image and warps the reference prediction accordingly.
The alignment module also estimates a pixel-wise warp confidence map $P_{\mathcal{R}}$.
(2) A non-parametric refinement module (Line~\ref{alglst:refine}) fuses the target and warped reference predictions\textemdash using $P_{\mathcal{R}}$ for the fusion weights\textemdash to produce refined target predictions.
The target predictions are subsequently converted to pseudo-labels according to the base UDA method (Line~\ref{alglst:pseudo}, \eg through $\argmax$ and confidence weighting if Refign is built on DACS~\cite{tranheden2021dacs}).
Since Refign hinges on high-quality reference predictions, we adapt $f_\theta$ to $\mathcal{R}$ in every second training iteration via the employed UDA base method (Lines~\ref{alglst:adr_start}-\ref{alglst:adr_end}, omitted in Fig.~\ref{fig:method_overview}).

Refign does not introduce any additional training parameters, since the alignment module is pre-trained and frozen, and the refinement module is non-parametric.
Consequently, the memory and computation overhead during training is minor since no additional backpropagation is required.
During inference, Refign is removed altogether.
We describe the main two components of Refign\textemdash the alignment and refinement modules\textemdash in more detail in Sec.~\ref{sec:method:alignment} and Sec.~\ref{sec:method:refinement} respectively.

\subsection{Alignment}
\label{sec:method:alignment}

Exact spatial alignment of the target and reference images is a crucial preliminary step for precise, pixel-wise semantic label refinement.
Our alignment module warps the reference image to align it with the target and estimates a confidence map of the warp, which is an important asset to guide the downstream label refinement.
To fulfill these requirements, we extend the warp consistency (WarpC) framework of~\cite{truong2021warp} with uncertainty prediction.

\begin{figure}
    \begin{minipage}[c]{0.49\linewidth}
    \includegraphics[width=\linewidth]{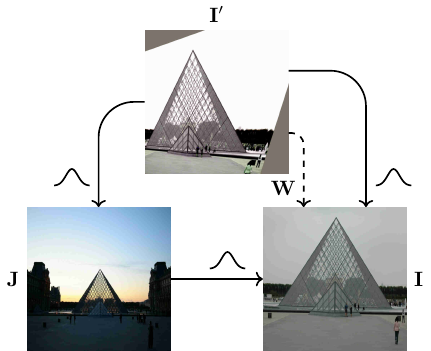} 
    \end{minipage}\hfill
    \begin{minipage}[c]{0.49\linewidth}
    \caption{Warp consistency graph of~\cite{truong2021warp}. Synthetic warp $\mathbf{W}$ supervises two flows: (1) the direct flow $\mathbf{I}'\rightarrow \mathbf{I}$, and (2) the composite flow $\mathbf{I}'\rightarrow \mathbf{J}\rightarrow \mathbf{I}$. We propose a probabilistic extension to~\cite{truong2021warp} by predicting parameterized flow distributions (shown as \raisebox{-0.1em}{\begin{tabular}{@{}c@{}}\includegraphics[width=3ex]{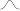}\end{tabular}}).}
    \label{fig:warp_consistency}
    \end{minipage}
\end{figure}

\PAR{WarpC.}
We first recap WarpC, referring to the original work~\cite{truong2021warp} for a more in-depth discussion.
Given two images $\mathbf{I}, \mathbf{J} \in \mathbb{R}^{h\times w\times 3}$ depicting a similar scene, the goal is to find a dense displacement field $\mathbf{F}_{\mathbf{J}\rightarrow \mathbf{I}} \in \mathbb{R}^{h\times w\times 2}$ relating pixels in $\mathbf{J}$ to $\mathbf{I}$. 
WarpC exploits the consistency graph shown in Fig.~\ref{fig:warp_consistency} to train the flow estimator. 
$\mathbf{I}$ is augmented heavily\textemdash\eg through a randomly sampled homography\textemdash to yield $\mathbf{I}'$. 
The synthetic augmentation warp $\mathbf{W}$ subsequently supervises two objectives: 
(1) the direct estimate $\hat{\mathbf{F}}_{\mathbf{I}'\rightarrow \mathbf{I}}$ of the flow $\mathbf{F}_{\mathbf{I}'\rightarrow \mathbf{I}}$,
\begin{equation}
    \label{eq:direct_flow}
    \mathcal{L}_{\mathbf{I}'\rightarrow \mathbf{I}} = \left\| \hat{\mathbf{F}}_{\mathbf{I}'\rightarrow \mathbf{I}} - \mathbf{W} \right\|^2,
\end{equation}
and (2) estimation of the composite flow $\mathbf{F}_{\mathbf{I}'\rightarrow \mathbf{J}\rightarrow \mathbf{I}}$ formed by chaining $\mathbf{F}_{\mathbf{I}'\rightarrow \mathbf{J}}$ and $\mathbf{F}_{\mathbf{J}\rightarrow \mathbf{I}}$:
\begin{equation}
    \begin{aligned}
        \mathcal{L}_{\mathbf{I}'\rightarrow \mathbf{J}\rightarrow \mathbf{I}} &= \left\| V \cdot \left( \hat{\mathbf{F}}_{\mathbf{I}'\rightarrow \mathbf{J}} + \Phi_{\hat{\mathbf{F}}_{\mathbf{I}'\rightarrow \mathbf{J}}}(\hat{\mathbf{F}}_{\mathbf{J}\rightarrow \mathbf{I}}) - \mathbf{W}\right) \right\|^2  \\
        &= \left\| V \cdot \left( \hat{\mathbf{F}}_{\mathbf{I}'\rightarrow \mathbf{J}\rightarrow \mathbf{I}} - \mathbf{W}\right) \right\|^2.
    \end{aligned}
    \label{eq:composite_flow}
\end{equation}
$\Phi_\mathbf{F}(\mathbf{T})$ defines the warp of $\mathbf{T}$ by the flow $\mathbf{F}$ and $V \in \{0,1\}^{h\times w}$ is the estimated visibility mask. $V$ aims to mask out all pixels in $\mathbf{I}'$ which have no correspondence in $\mathbf{J}$ due to occlusion, image boundary, \etc. We estimate $V$ analogously to~\cite{truong2021warp} based on the Cauchy–Schwarz inequality (see appendix, Sec.~\ref{subsec:alignment_training_details}).
The two loss terms in \eqref{eq:direct_flow} and \eqref{eq:composite_flow} complement each other: $\mathcal{L}_{\mathbf{I}'\rightarrow \mathbf{I}}$ promotes convergence and favors smooth solutions, while $\mathcal{L}_{\mathbf{I}'\rightarrow \mathbf{J}\rightarrow \mathbf{I}}$ learns realistic motion patterns and appearance changes.
The overall network is trained via $\mathcal{L}_{\text{align}} = \mathcal{L}_{\mathbf{I}'\rightarrow \mathbf{I}} + \lambda \mathcal{L}_{\mathbf{I}'\rightarrow \mathbf{J}\rightarrow \mathbf{I}}$, where $\lambda$ is a weighting term balancing the individual losses.

\PAR{UAWarpC.}
Our extension, Uncertainty-Aware WarpC (UAWarpC) adds predictive uncertainty estimation~\cite{nix1994estimating} to WarpC.
We model the flow conditioned on the image inputs $\mathbf{I}$, $\mathbf{J}$ via a Gaussian $p(\mathbf{F}_{\mathbf{J}\rightarrow \mathbf{I}} | \mathbf{I}, \mathbf{J}) = \mathcal{N}(\mathbf{F}_{\mathbf{J}\rightarrow \mathbf{I}}; \hat{\mathbf{F}}_{\mathbf{J}\rightarrow \mathbf{I}}, \hat{\Sigma}_{\mathbf{J}\rightarrow \mathbf{I}})$, implying that the predicted flow is corrupted with additive Gaussian noise.
Note that a different Gaussian is predicted for each pixel.
To accommodate $x$ and $y$ flow directions, the distributions are bivariate.
We assume for simplicity that the variance is equal in both directions.
Thus, the network is trained to output mean $\hat{\mathbf{F}}^{ij} \in \mathbb{R}^2$ and log-variance $\log\hat{\Sigma}^{ij} \in \mathbb{R}$ at each spatial location $ij$. 

Estimating the distribution of the composite flow $\mathbf{F}_{\mathbf{I}'\rightarrow \mathbf{J}\rightarrow \mathbf{I}}$ precisely would be computationally infeasible in our case.
Instead, we take the simplifying assumption that the flow predictions $\mathbf{F}_{\mathbf{I}'\rightarrow \mathbf{J}}$ and $\mathbf{F}_{\mathbf{J}\rightarrow \mathbf{I}}$ are conditionally independent random variables given the images.
Their sum is normally distributed with a mean and variance equal to the sum of the two means and variances.
We thus model 
$p(\mathbf{F}_{\mathbf{I}'\rightarrow \mathbf{J}\rightarrow \mathbf{I}} | \mathbf{I}, \mathbf{J}, \mathbf{I}') = \mathcal{N}(\mathbf{F}_{\mathbf{I}'\rightarrow \mathbf{J}\rightarrow \mathbf{I}}; \hat{\mathbf{F}}_{\mathbf{I}'\rightarrow \mathbf{J}\rightarrow\mathbf{I}}, \hat{\Sigma}_{\mathbf{I'}\rightarrow \mathbf{J}\rightarrow\mathbf{I}})$
as another Gaussian.
Analogously to the composite flow mean in \eqref{eq:composite_flow}, the composite flow variance is calculated through warping.
\begin{equation}
    \hat{\Sigma}_{\mathbf{I'}\rightarrow \mathbf{J}\rightarrow\mathbf{I}} = \hat{\Sigma}_{\mathbf{I'}\rightarrow \mathbf{J}} + \Phi_{\hat{\mathbf{F}}_{\mathbf{I'}\rightarrow \mathbf{J}}}(\hat{\Sigma}_{\mathbf{J}\rightarrow\mathbf{I}})
\end{equation}
We follow the principle of maximum log-likelihood estimation to train our model (derivation in appendix, Sec.~\ref{sec:extra_derivations}).
\begin{equation}
    \begin{aligned}
        \mathcal{L}^{\text{prob}}_{\mathbf{I'}\rightarrow \mathbf{I}} &= -\log p(\mathbf{W} | \mathbf{I}, \mathbf{I'}) \\
        &\propto  \frac{1}{2\hat{\Sigma}_{\mathbf{I'}\rightarrow \mathbf{I}}} \mathcal{L}_{\mathbf{I'}\rightarrow \mathbf{I}} + \log \hat{\Sigma}_{\mathbf{I'}\rightarrow \mathbf{I}}
    \end{aligned}
\end{equation}
The formulation for $\mathcal{L}^{\text{prob}}_{\mathbf{I'}\rightarrow \mathbf{J}\rightarrow\mathbf{I}}$ is obtained simply by replacing the subscripts.
Although the negative log-likelihood of a Gaussian corresponds to the squared error loss, in practice we use a Huber loss~\cite{huber1992robust} in \eqref{eq:direct_flow} and \eqref{eq:composite_flow} to increase robustness to outliers.

\PAR{Alignment for Label Refinement.}
The alignment module is trained separately on the large-scale MegaDepth~\cite{li2018megadepth} dataset and it is subsequently frozen during self-training of the segmentation network.
During self-training, it estimates the flow $\mathbf{F}_{\mathbf{I}_\mathcal{T}\rightarrow\mathbf{I}_\mathcal{R}}$ and accordingly warps the reference class-wise probability map $\mathbf{Q}_{\mathcal{R}}\in \mathbb{R}^{h\times w\times c}$, yielding $\mathbf{Q}^a_{\mathcal{R}}$ (see Fig.~\ref{fig:method_overview}).
In addition, it estimates a warp confidence map $P_{\mathcal{R}}\in [0, 1]^{h\times w}$.
To obtain $P_{\mathcal{R}}$ from our probabilistic model, we compute the probability of the true flow $\mathbf{F}_{\mathbf{I}_\mathcal{T}\rightarrow\mathbf{I}_\mathcal{R}}$ being within a radius $r$ of the estimated flow $\hat{\mathbf{F}}_{\mathbf{I}_\mathcal{T}\rightarrow\mathbf{I}_\mathcal{R}}$, as in~\cite{truong2021pdc} (derivation in appendix, Sec.~\ref{sec:extra_derivations}).
\begin{equation}
    P_{\mathcal{R}} = p(\| \mathbf{F}_{\mathbf{I}_\mathcal{T}\rightarrow\mathbf{I}_\mathcal{R}} - \hat{\mathbf{F}}_{\mathbf{I}_\mathcal{T}\rightarrow\mathbf{I}_\mathcal{R}} \| \leq r) = 1 - \exp{\frac{-r^2}{2 \hat{\Sigma}_{\mathbf{I}_\mathcal{T}\rightarrow\mathbf{I}_\mathcal{R}}}}
\end{equation}
We set $r=1$. The elements of $P_{\mathcal{R}}$ corresponding to invalid warp regions are set to zero.

\subsection{Refinement}
\label{sec:method:refinement}

The refinement module aims to improve target class-wise probabilities $\mathbf{Q}_{\mathcal{T}}$ using the aligned reference class probabilities $\mathbf{Q}^a_{\mathcal{R}}$ and the matching confidence map $P_{\mathcal{R}}$.
The refined target class weights $\mathbf{Q}^r_{\mathcal{T}}$ are then converted to pseudo-labels for self-training.
The refinement is a convex combination with element-wise weights $\alpha \in \mathbb{R}^{h\times w\times c}$:
\begin{equation}
    \mathbf{Q}^r_{\mathcal{T}} = (1 - \alpha) \odot \mathbf{Q}_{\mathcal{T}} + \alpha \odot \mathbf{Q}^a_{\mathcal{R}},
    \label{eq:convex_comb}
\end{equation}
where $\odot$ denotes element-wise multiplication. Our construction of $\alpha$ builds on principles of self-label correction~\cite{wang2021proselflc}, as we describe in the following.

\PAR{Confidence.}
During early training stages, the network's predictions are unreliable, especially in the more challenging adverse domain.
In line with the principle of curriculum learning~\cite{bengio2009curriculum}, the model should thus rely more heavily on the ``easier'' reference images first.
Even if the reference prediction guidance is inaccurate\textemdash \eg due to erroneous warping\textemdash degradation during early training is limited, as deep networks tend to learn simple patterns first before memorizing noise~\cite{arpit2017closer}.
Later in training, on the other hand, the model should be allowed to ignore or revise faulty reference predictions.
This progression can be captured via the model confidence, which increases steadily during training.
More formally, we gauge the model confidence with the normalized entropy of the target probability map $\hat{H}(\mathbf{Q}_{\mathcal{T}}) = \frac{H(\mathbf{Q}_{\mathcal{T}})}{H_{\text{max}}} \in [0,1]^{h\times w}$.
We take the mean over all pixels to obtain a global image-level estimate, and introduce a hyperparameter $\gamma$ as an exponent to allow for tuning.
This yields a trust score $s$:
\begin{equation}
    \alpha \propto s(\mathbf{Q}_{\mathcal{T}}) = \left(\mean\left\{\hat{H}(\mathbf{Q}_{\mathcal{T}})\right\}\right)^\gamma.
    \label{eq:gamma_eq}
\end{equation}

\PAR{Large Static Classes.}
\looseness=-1  % hacky
Based on the average size of connected class segments, we hereafter refer to the three classes \emph{pole}, \emph{traffic light}, and \emph{traffic sign} as \emph{small static classes} (avg.\ size of 8k pixels on Cityscapes~\cite{cordts2016cityscapes}), while the other eight static classes are named \emph{large static classes} (avg.\ size of 234k pixels).
We experimentally observe that large static classes are more accurately matched by the alignment module compared to small static classes (see appendix, Sec.~\ref{sec:large_static_extra}).
In fact, guiding the refinement through $P_{\mathcal{R}}$ for large static classes might be overly pessimistic.
The matching network learns to be uncertain in non-textured regions (\eg \emph{road}, \emph{sky}), where it is unable to identify distinct matches~\cite{truong2021pdc}.
However, even though $P_{\mathcal{R}}$ is low for these regions, the broader semantic class is still matched correctly, due to smooth interpolation learned by the alignment network.

We propose more aggressive refinement for large static classes to compensate for this effect.
To rule out unwanted drift towards large classes, we restrict the aggressive mixing both \emph{spatially} and \emph{across channels} via a binary mask $\mathbf{M} \in \{0, 1\}^{h\times w\times c}$ with elements $m^{ijk}$.
We define $\mathcal{A}$ as the set of large static classes, $Z_{\mathcal{T}} = \argmax_c \mathbf{Q}_{\mathcal{T}}$ as the target predictions, and $Z_{\mathcal{R}}$ as the reference predictions.
\begin{equation}
    m^{ijk} = \begin{cases}1~&{\text{ if }}~k\in \mathcal{A}\text{ and } Z_{\mathcal{T}}^{ij}\in \mathcal{A}\text{ and }Z_{\mathcal{R}}^{ij}\in \mathcal{A}~,\\0~&{\text{ otherwise }}.\end{cases}
\end{equation}
$\mathbf{M}$ restricts aggressive mixing only to tensor elements fulfilling two criteria:
(1) The element belongs to the channel of a large static class.
(2) The corresponding pixel is labeled with a large static class in \emph{both} domains.
Aggressive mixing is incorporated as follows:
\begin{equation}
    \alpha \propto \max (P_{\mathcal{R}}, \mathbf{M}).
\end{equation}
As slight abuse of notation, we use $\max(\cdot, \cdot)$ to denote the element-wise maximum of two tensors, which are broadcast to the same shape.
Here, $P_\mathcal{R}$ is stacked $c$ times along the third dimension to match the shape of $\mathbf{M}$.

\PAR{Refinement Equation.}
Combining the two propositions, we obtain adaptive pseudo-label refinement:
\begin{equation}
    \alpha = s(\mathbf{Q}_{\mathcal{T}}) \max (P_{\mathcal{R}}, \mathbf{M}).
    \label{eq:epsilon_2}
\end{equation}
Owing to pixel-wise modulation by $P_{\mathcal{R}}$, this refinement scheme can ignore dynamic objects and small static objects, which are difficult to align.
On the other hand, if \eg two cars are coincidentally present at the same location, information transfer is still possible.
Furthermore, the scheme allows for easy tuning of the degree of mixing with a single hyperparameter\textemdash the exponent $\gamma$ of trust score $s$.
Finally, since entries of $P_{\mathcal{R}}$ corresponding to invalid warp regions are zero, no mixing happens if no match can be found.

\begin{table*}
\caption{Comparison to the state of the art in Cityscapes\textrightarrow ACDC domain adaptation on the ACDC test set. Methods above the double line use a DeepLabv2 model. ``Ref.'': for each adverse input image a reference frame from the same geo-location is used.}%
\smallskip%
\centering%
\resizebox{\linewidth}{!}{%
\ra{1.3}%
\begin{tabular}{@{}lcccccccccccccccccccccccr@{}}\toprule
\multirow{2}{*}{Method} && \multirow{2}{*}{Ref.} && \multicolumn{21}{c}{IoU\,$\uparrow$} \\
\cmidrule{5-25} &&&& \rotatebox[origin=c]{90}{road} & \rotatebox[origin=c]{90}{sidew.} & \rotatebox[origin=c]{90}{build.} & \rotatebox[origin=c]{90}{wall} & \rotatebox[origin=c]{90}{fence} & \rotatebox[origin=c]{90}{pole} & \rotatebox[origin=c]{90}{light} & \rotatebox[origin=c]{90}{sign} & \rotatebox[origin=c]{90}{veget.} & \rotatebox[origin=c]{90}{terrain} & \rotatebox[origin=c]{90}{sky} & \rotatebox[origin=c]{90}{person} & \rotatebox[origin=c]{90}{rider} & \rotatebox[origin=c]{90}{car} & \rotatebox[origin=c]{90}{truck} & \rotatebox[origin=c]{90}{bus} & \rotatebox[origin=c]{90}{train} & \rotatebox[origin=c]{90}{motorc.} & \rotatebox[origin=c]{90}{bicycle} && \multicolumn{1}{c}{\phantom{00}\rotatebox[origin=c]{90}{\textbf{mean}}} \\ \midrule
% DeepLabv2~\cite{chen2017deeplab} &&&& 71.9 & 26.2 & 51.1 & 18.8 & 22.5 & 19.7 & 33.0 & 27.7 & 67.9 & 28.6 & 44.2 & 43.1 & 22.1 & 71.2 & 29.8 & 33.3 & 48.4 & 26.2 & 35.8 && 38.0 \\
% Oracle* && 88.0 & 62.3 & 80.8 & 37.0 & 35.1 & 33.9 & 49.8 & 49.5 & 80.1 & 50.7 & 92.5 & 51.1 & 26.5 & 79.9 & 49.0 & 41.1 & 72.2 & 26.5 & 44.2 && 55.3 \\ 
% ADVENT~\cite{vu2019advent} &&&&  72.9 & 14.3 & 40.5 & 16.6 & 21.2 & \phantom{0}9.3 & 17.4 & 21.2 & 63.8 & 23.8 & 18.3 & 32.6 & 19.5 & 69.5 & 36.2 & 34.5 & 46.2 & 26.9 & 36.1 && 32.7 \\
% CRST~\cite{zou2019confidence} &&&& 51.7 & 24.4 & 67.8 & 13.3 & \phantom{0}9.7 & 30.2 & 38.2 & 34.1 & 58.0 & 25.2 & 76.8 & 39.9 & 17.1 & 65.4 & \phantom{0}3.7 & \phantom{0}6.6 & 39.6 & 11.8 & \phantom{0}8.6 && 32.8 \\
AdaptSegNet~\cite{tsai2018learning} &&&& 69.4 & 34.0 & 52.8 & 13.5 & 18.0 & \phantom{0}4.3 & 14.9 & \phantom{0}9.7 & 64.0 & 23.1 & 38.2 & 38.6 & 20.1 & 59.3 & 35.6 & 30.6 & 53.9 & 19.8 & 33.9 && 33.4 \\
% SIM~\cite{wang2020differential} &&&& 53.8 & \phantom{0}6.8 & 75.5 & 11.6 & 22.3 & 11.7 & 23.4 & 25.7 & 66.1 & \phantom{0}8.3 & 80.6 & 41.8 & 24.8 & 49.7 & 38.6 & 21.0 & 41.8 & 25.1 & 29.6 && 34.6 \\
% MRNet~\cite{zheng2021rectifying} &&&& 72.2 & \phantom{0}8.2 & 36.4 & 13.7 & 18.5 & 20.4 & 38.7 & 45.4 & 70.2 & 35.7 & \phantom{0}5.0 & 47.8 & 19.1 & 73.6 & 42.1 & 36.0 & 47.4 & 17.7 & 37.4 && 36.1 \\
BDL~\cite{li2019bidirectional} &&&& 56.0 & 32.5 & 68.1 & 20.1 & 17.4 & 15.8 & 30.2 & 28.7 & 59.9 & 25.3 & 37.7 & 28.7 & 25.5 & 70.2 & 39.6 & 40.5 & 52.7 & 29.2 & 38.4 && 37.7 \\
% CLAN~\cite{luo2019taking} &&&& 79.1 & 29.5 & 45.9 & 18.1 & 21.3 & 22.1 & 35.3 & 40.7 & 67.4 & 29.4 & 32.8 & 42.7 & 18.5 & 73.6 & 42.0 & 31.6 & 55.7 & 25.4 & 30.7 && 39.0 \\
FDA~\cite{yang2020fda} &&&& 73.2 & 34.7 & 59.0 & 24.8 & 29.5 & 28.6 & 43.3 & 44.9 & 70.1 & 28.2 & 54.7 & 47.0 & 28.5 & 74.6 & 44.8 & 52.3 & 63.3 & 28.3 & 39.5 && 45.7 \\
DANNet (DeepLabv2)~\cite{wu2021dannet} && \checkmark && 82.9 & 53.1 & 75.3 & 32.1 & 28.2 & 26.5 & 39.4 & 40.3 & 70.0 & 39.7 & 83.5 & 42.8 & 28.9 & 68.0 & 32.0 & 31.6 & 47.0 & 21.5 & 36.7 && 46.3 \\
DANIA (DeepLabv2)~\cite{wu2021one} && \checkmark && 87.8 & 57.1 & 80.3 & 36.2 & 31.4 & 28.6 & 49.5 & 45.8 & 76.2 & 48.8 & 90.2 & 47.9 & 31.1 & 75.5 & 36.5 & 36.5 & 47.8 & 32.5 & 44.1 && 51.8 \\
\midrule
DACS~\cite{tranheden2021dacs} &&&& 58.5 & 34.7 & 76.4 & 20.9 & 22.6 & 31.7 & 32.7 & 46.8 & 58.7 & 39.0 & 36.3 & 43.7 & 20.5 & 72.3 & 39.6 & 34.8 & 51.1 & 24.6 & 38.2 && 41.2 \\
% DACS w/ Test-Refign && \checkmark && 60.7 & 38.1 & 77.8 & 22.1 & 25.2 & 32.0 & 33.1 & 47.5 & 60.8 & 41.6 & 39.9 & 43.6 & 20.5 & 72.8 & 39.4 & 35.0 & 51.7 & 24.5 & 38.6 && 42.4 \\
Refign-DACS (ours) && \checkmark && 49.5 & 56.7 & 79.8 & 31.2 & 25.7 & 34.1 & 48.0 & 48.7 & 76.2 & 42.5 & 38.5 & 48.3 & 24.7 & 75.3 & 46.5 & 43.9 & 64.3 & 34.1 & 43.6 && 48.0 \\
% \midrule
% FDA w/ Test-Refign && \checkmark && 80.4 & 37.2 & 60.7 & 27.9 & 30.7 & 30.1 & 45.0 & 46.5 & 72.3 & 32.7 & 61.3 & 47.3 & 30.2 & 76.2 & 47.2 & 50.0 & 64.5 & 25.9 & 41.4 && 47.7 \\
% Refign-FDA && \checkmark && 79.9 & 37.7 & 62.4 & 26.2 & 28.8 & 29.4 & 44.5 & 44.8 & 72.0 & 29.6 & 65.9 & 45.8 & 28.1 & 73.5 & 42.5 & 45.7 & 62.0 & 24.8 & 39.6 && 46.5 \\
\midrule\midrule
% GCMA (RefineNet)~\cite{sakaridis2019guided} && \checkmark && 79.7 & 48.7 & 71.5 & 21.6 & 29.9 & 42.5 & 56.7 & 57.7 & 75.8 & 39.5 & 87.2 & 57.4 & 29.7 & 80.6 & 44.9 & 46.2 & 62.0 & 37.2 & 46.5 && 53.4 \\
MGCDA (RefineNet)~\cite{sakaridis2020map} && \checkmark && 73.4 & 28.7 & 69.9 & 19.3 & 26.3 & 36.8 & 53.0 & 53.3 & 75.4 & 32.0 & 84.6 & 51.0 & 26.1 & 77.6 & 43.2 & 45.9 & 53.9 & 32.7 & 41.5 && 48.7 \\
DANNet (PSPNet)~\cite{wu2021dannet} && \checkmark && 84.3 & 54.2 & 77.6 & 38.0 & 30.0 & 18.9 & 41.6 & 35.2 & 71.3 & 39.4 & 86.6 & 48.7 & 29.2 & 76.2 & 41.6 & 43.0 & 58.6 & 32.6 & 43.9 && 50.0 \\
DANIA (PSPNet)~\cite{wu2021one} && \checkmark && 88.4 & 60.6 & 81.1 & 37.1 & 32.8 & 28.4 & 43.2 & 42.6 & 77.7 & 50.5 & 90.5 & 51.5 & 31.1 & 76.0 & 37.4 & 44.9 & 64.0 & 31.8 & 46.3 && 53.5 \\
\midrule
DAFormer~\cite{hoyer2021daformer} &&&& 58.4 & 51.3 & 84.0 & 42.7 & \textbf{35.1} & 50.7 & 30.0 & 57.0 & 74.8 & 52.8 & 51.3 & 58.3 & 32.6 & 82.7 & 58.3 & 54.9 & 82.4 & 44.1 & 50.7 && 55.4 \\
% DAFormer w/ Test-Refign && \checkmark && 58.1 & 46.7 & 85.7 & \textbf{48.2} & \textbf{36.8} & 48.9 & 46.1 & 58.2 & 71.1 & 54.2 & 48.8 & 56.7 & 28.2 & 82.9 & 57.7 & 64.2 & 84.1 & 40.4 & 51.0 && 56.2 \\
Refign-DAFormer (ours) && \checkmark && \textbf{89.5} & \textbf{63.4} & \textbf{87.3} & \textbf{43.6} & 34.3 & \textbf{52.3} & \textbf{63.2} & \textbf{61.4} & \textbf{86.9} & \textbf{58.5} & \textbf{95.7} & \textbf{62.1} & \textbf{39.3} & \textbf{84.1} & \textbf{65.7} & \textbf{71.3} & \textbf{85.4} & \textbf{47.9} & \textbf{52.8} && \textbf{65.5} \\
\bottomrule
\end{tabular}%
}
\label{tab:acdc_sota}
\vspace{-3mm}
\end{table*}

\begin{table}
\caption{Comparison of Cityscapes\textrightarrow Dark Zurich methods on Dark Zurich-test. Trained models are tested for generalization on the Nighttime Driving (ND) and BDD100k-night (Bn) test sets.}%
\smallskip%
\centering%
\resizebox{\columnwidth}{!}{%
\ra{1.3}%
\setlength{\tabcolsep}{12pt}%
\begin{tabular}{@{}lcccr@{}}\toprule
\multirow{2}{*}{Method} && \multicolumn{3}{c}{mIoU\,$\uparrow$} \\
\cmidrule{3-5} && Dark Zurich-test~\cite{sakaridis2020map} & ND~\cite{dai2018dark} & Bn~\cite{yu2020bdd100k,sakaridis2020map}\\
\midrule
% AdaptSegNet (DeepLabv2)~\cite{tsai2018learning} && 30.4 & 34.5 & 22.0 \\
% ADVENT (DeepLabv2)~\cite{vu2019advent} && 29.7 & 34.7 & 22.6 \\
% BDL (DeepLabv2)~\cite{li2019bidirectional} && 30.8 & 34.7 & 22.8 \\
DMAda (RefineNet)~\cite{dai2018dark} && 32.1 & 36.1 & 28.3 \\
GCMA (RefineNet)~\cite{sakaridis2019guided} && 42.0 & 45.6 & 33.2 \\
MGCDA (RefineNet)~\cite{sakaridis2020map} && 42.5 & 49.4 & 34.9 \\
CDAda (RefineNet)~\cite{xu2021cdada} && 45.0 & 50.9 & 33.8 \\
DANNet (PSPNet)~\cite{wu2021dannet} && 45.2 & 47.7 & 28.0 \\
% Bi-Mix (RefineNet)~\cite{yang2021bi} && 46.5 & 51.6 & 24.9 \\ % they overfit to different benchmarks
DANIA (PSPNet)~\cite{wu2021one} && 47.0 & 48.4 & 27.0 \\
CCDistill (RefineNet)~\cite{gao2022cross} && 47.5 & 46.2 & 33.0 \\
% SePiCo (DAFormer)~\cite{xie2022sepico} && 54.2 & \textbf{57.1} & \textbf{36.9} \\
\midrule
DACS (DeepLabv2)~\cite{tranheden2021dacs} && 36.7 & 39.5 & 25.3 \\
Refign-DACS (DeepLabv2, ours) && 41.2 & 41.5 & 26.2 \\
\midrule
DAFormer~\cite{hoyer2021daformer} && 53.8 & 54.1 & 33.8 \\
Refign-DAFormer (ours) && \textbf{56.2} & \textbf{56.8} & \textbf{35.2} \\
\bottomrule
\end{tabular}%
}
\label{tab:night_generalization}
\vspace{-3mm}
\end{table}

%%%%%%%%% EXPERIMENTS
\section{Experiments}
\label{sec:exp}

We present extensive experiments for both UDA and geometric matching.
Sec.~\ref{sec:exp:setup} provides an overview of the experimental setup.
Sec.~\ref{sec:exp:sota_uda} and Sec.~\ref{sec:exp:robotcar_cmu} present comparisons with state-of-the-art methods in UDA and semi-supervised domain adaptation, respectively.
Sec.~\ref{sec:exp:ablations} discusses ablations and Sec.~\ref{sec:exp:matching} shows geometric matching comparisons.
Training settings and implementation details are discussed in Sec.~\ref{sec:training_details} of the appendix.

\begin{table}
\caption{Semi-supervised domain adaptation on Cityscapes\textrightarrow RobotCar and Cityscapes\textrightarrow CMU.
``Ref.'': a reference frame is used for each adverse input image.}%
\smallskip%
\centering%
\resizebox{\linewidth}{!}{%
\ra{1.3}%
\setlength{\tabcolsep}{9pt}%
\begin{tabular}{@{}lccccr@{}}\toprule
\multirow{2}{*}{Method} && \multirow{2}{*}{Ref.} && \multicolumn{2}{c}{mIoU\,$\uparrow$} \\
\cmidrule{5-6} &&&& RobotCar~\cite{maddern20171,larsson2019cross} & CMU~\cite{badino2011visual,larsson2019cross}\\
\midrule
% PSPNet~\cite{zhao2017pyramid} &&&&& 22.2 & 29.9  \\
PSPNet~\cite{zhao2017pyramid} &&&& 45.8 & 73.6 \\
Cross-Season, CE~\cite{larsson2019cross} && \checkmark && 53.8 & 79.3 \\
Cross-Season, Hinge\textsubscript{\textit{C}}~\cite{larsson2019cross} && \checkmark && 50.6 & 72.4 \\
Cross-Season, Hinge\textsubscript{\textit{F}}~\cite{larsson2019cross} && \checkmark && 55.4 & 75.3 \\
\midrule
DAFormer~\cite{hoyer2021daformer} &&&& 51.7 & 75.6 \\
Refign-DAFormer (ours) && \checkmark && \textbf{60.5} & \textbf{83.6} \\
\bottomrule
\end{tabular}
%
}
\label{tab:robotcar_cmu}
\vspace{-5mm}
\end{table}

\subsection{Setup}
\label{sec:exp:setup}

\PAR{Datasets.}
For the source domain we use Cityscapes~\cite{cordts2016cityscapes}.
For the target and reference domains, we use ACDC~\cite{sakaridis2021acdc}, Dark Zurich~\cite{sakaridis2020map}, RobotCar Correspondence~\cite{maddern20171,larsson2019cross}, or CMU Correspondence~\cite{badino2011visual,larsson2019cross}.
Each of these four target-domain datasets contains adverse-normal condition street scene image pairs in the training set.
ACDC contains 1600 training, 406 validation, and 2000 test images distributed equally among fog, night, rain, and snow.
Dark Zurich contains 2416 training, 50 validation, and 151 test images for nighttime.
RobotCar (resp.\ CMU) Correspondence contains 6511 (28766) training, 27 (25) validation, and 27 (33) test images, captured at various conditions.
The RobotCar and CMU Correspondence datasets additionally have 40 and 66 coarsely annotated images, enabling semi-supervised domain adaptation.
For training the alignment network, we use MegaDepth~\cite{li2018megadepth} and evaluate using the test split of~\cite{shen2020ransac}.
To test the ability of the alignment module to generalize to road scenes, we additionally evaluate it on the sparse ground-truth matches provided by~\cite{larsson2019cross} for the RobotCar and CMU Correspondence datasets.

\PAR{Architectures.}
To showcase the flexibility of Refign, we combine it with state-of-the-art UDA methods.
We choose DACS~\cite{tranheden2021dacs} (using DeepLabv2~\cite{chen2017deeplab}) and DAFormer~\cite{hoyer2021daformer} (based on SegFormer~\cite{xie2021segformer}) as base methods.
Our alignment network follows almost exactly the same architecture as WarpC~\cite{truong2021warp} (VGG-16~\cite{simonyan2014very} encoder and GLU-Net~\cite{truong2020glu} decoder), complemented with the uncertainty decoder of~\cite{truong2021pdc}.

\PAR{Metrics.}
For evaluating segmentation results, we use mean intersection over union (mIoU).
Geometric matching accuracy is evaluated using the percentage of correct keypoints at a given pixel threshold $T$ (PCK-$T$).
The quality of matching uncertainty estimates is evaluated using sparsification error, specifically the area under the sparsification error curve (AUSE)~\cite{truong2021pdc} for average end-point error (AEPE).

\begin{figure*}
    \adjustbox{max width=\textwidth}{%
        \input{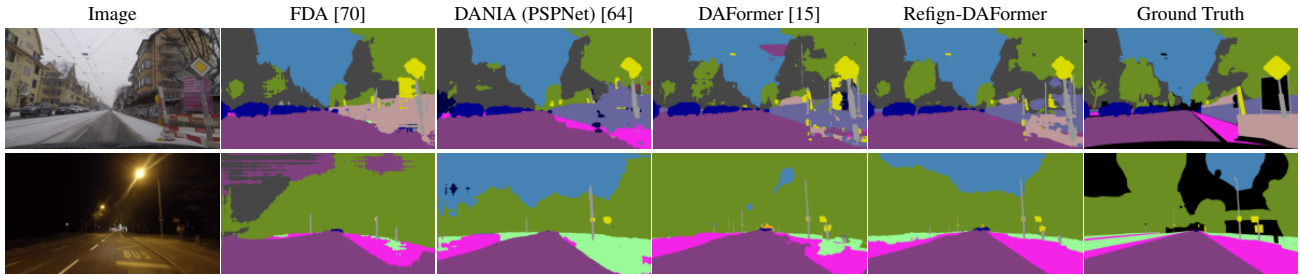}%
    }%
    \caption{Qualitative segmentation results on two images of the ACDC validation set for models adapted from Cityscapes to ACDC.}
    \label{fig:qualitative_seg}
\end{figure*}

\subsection{Comparison to the State of the Art in UDA}
\label{sec:exp:sota_uda}

\begin{table*}
\caption{Ablation study on the ACDC validation set (metric: IoU) for different components of Refign, as detailed in (\ref{eq:epsilon_2}).
Default values in case of component omission are: $P_\mathcal{R}=\frac{1}{2}$, $\mathbf{M}=0$, $s=1$.
``$\mathcal{R}$-ad'': concurrent adaptation to $\mathcal{R}$, \ie, Lines~\ref{alglst:adr_start}-\ref{alglst:adr_end} of Alg.~\ref{alg:refign}.}%
\smallskip%
\centering%
\resizebox{\linewidth}{!}{%
    \ra{1.3}%
%
% \documentclass[11pt]{article}  % for debug
% \usepackage{booktabs}
% \usepackage{multirow}
% \newcommand{\ra}[1]{\renewcommand{\arraystretch}{#1}}
% \usepackage{amsmath}
% \usepackage{amssymb}
% \usepackage{adjustbox}
%
% see here http://www.traag.net/2014/06/05/281/
\pgfplotsset{%
    colormap/RdBu-5,%  % lower number --> lighter
}%
\pgfplotstableset{%
    % multicolumn names={c},
    header=false,
    col sep=&,
    row sep=\\,
    every head row/.style={output empty row},  
    every first row/.style={before row=\midrule, after row=\midrule},
    every row no 4/.style={after row=\midrule},
    every last row/.style={after row=\bottomrule},
    /color cells/min/.initial=0,
    /color cells/max/.initial=1000,
    /color cells/textcolor/.initial=,
    %
    % Usage: 'color cells={min=<value which is mapped to lowest color>, 
    %   max = <value which is mapped to largest>}
    color cells/.code={%
        \pgfqkeys{/color cells}{#1}%
        \pgfkeysalso{%
            postproc cell content/.code={%
                \begingroup
                %
                % acquire the value before any number printer changed
                % it:
                \pgfkeysgetvalue{/pgfplots/table/@preprocessed cell content}\value
                \ifx\value\empty
                    \endgroup
                \else
                \pgfmathfloatparsenumber{\value}%
                \pgfmathfloattofixed{\pgfmathresult}%
                \let\value=\pgfmathresult
                %
                % map that value:
                \pgfplotscolormapaccess
                    [\pgfkeysvalueof{/color cells/min}:\pgfkeysvalueof{/color cells/max}]
                    {\value}
                    {\pgfkeysvalueof{/pgfplots/colormap name}}%
                % now, \pgfmathresult contains {<R>,<G>,<B>}
                % 
                % acquire the value AFTER any preprocessor or
                % typesetter (like number printer) worked on it:
                \pgfkeysgetvalue{/pgfplots/table/@cell content}\typesetvalue
                \pgfkeysgetvalue{/color cells/textcolor}\textcolorvalue
                %
                % tex-expansion control
                % see https://tex.stackexchange.com/questions/12668/where-do-i-start-latex-programming/27589#27589
                \toks0=\expandafter{\typesetvalue}%
                \xdef\temp{%
                    \noexpand\pgfkeysalso{%
                        @cell content={%
                            \noexpand\cellcolor[rgb]{\pgfmathresult}%
                            \noexpand\definecolor{mapped color}{rgb}{\pgfmathresult}%
                            \ifx\textcolorvalue\empty
                            \else
                                \noexpand\color{\textcolorvalue}%
                            \fi
                            \the\toks0 %
                        }%
                    }%
                }%
                \endgroup
                \temp
                \fi
            }%
        }%
    }
}%
\def\mycr{77.7}% string which defines color width
\pgfmathsetmacro{\roadmin}{64.3 - \mycr}%
\pgfmathsetmacro{\roadmax}{64.3 + \mycr}%
\pgfmathsetmacro{\sidewmin}{25.4 - \mycr}%
\pgfmathsetmacro{\sidewmax}{25.4 + \mycr}%
\pgfmathsetmacro{\buildmin}{63.3 - \mycr}%
\pgfmathsetmacro{\buildmax}{63.3 + \mycr}%
\pgfmathsetmacro{\wallmin}{26.5 - \mycr}%
\pgfmathsetmacro{\wallmax}{26.5 + \mycr}%
\pgfmathsetmacro{\fencemin}{18.0 - \mycr}%
\pgfmathsetmacro{\fencemax}{18.0 + \mycr}%
\pgfmathsetmacro{\polemin}{5.5 - \mycr}%
\pgfmathsetmacro{\polemax}{5.5 + \mycr}%
\pgfmathsetmacro{\lightmin}{6.4 - \mycr}%
\pgfmathsetmacro{\lightmax}{6.4 + \mycr}%
\pgfmathsetmacro{\signmin}{9.3 - \mycr}%
\pgfmathsetmacro{\signmax}{9.3 + \mycr}%
\pgfmathsetmacro{\vegetmin}{63.7 - \mycr}%
\pgfmathsetmacro{\vegetmax}{63.7 + \mycr}%
\pgfmathsetmacro{\terrainmin}{13.3 - \mycr}%
\pgfmathsetmacro{\terrainmax}{13.3 + \mycr}%
\pgfmathsetmacro{\skymin}{79.1 - \mycr}%
\pgfmathsetmacro{\skymax}{79.1 + \mycr}%
\pgfmathsetmacro{\personmin}{6.9 - \mycr}%
\pgfmathsetmacro{\personmax}{6.9 + \mycr}%
\pgfmathsetmacro{\ridermin}{0.8 - \mycr}%
\pgfmathsetmacro{\ridermax}{0.8 + \mycr}%
\pgfmathsetmacro{\carmin}{23.8 - \mycr}%
\pgfmathsetmacro{\carmax}{23.8 + \mycr}%
\pgfmathsetmacro{\truckmin}{24.8 - \mycr}%
\pgfmathsetmacro{\truckmax}{24.8 + \mycr}%
\pgfmathsetmacro{\busmin}{39.5 - \mycr}%
\pgfmathsetmacro{\busmax}{39.5 + \mycr}%
\pgfmathsetmacro{\trainmin}{8.3 - \mycr}%
\pgfmathsetmacro{\trainmax}{8.3 + \mycr}%
\pgfmathsetmacro{\motorcmin}{6.7 - \mycr}%
\pgfmathsetmacro{\motorcmax}{6.7 + \mycr}%
\pgfmathsetmacro{\bicyclemin}{23.7 - \mycr}%
\pgfmathsetmacro{\bicyclemax}{23.7 + \mycr}%
\pgfmathsetmacro{\meanmin}{26.8 - \mycr}%
\pgfmathsetmacro{\meanmax}{26.8 + \mycr}%
% \pgfkeys{
%     /pgfplots/table/string type in dec sep align/.style={
%         string type,
%         postproc cell content/.code={%
%             \ifnum\pgfplotstablepartno=0%
%                 \pgfkeys{/pgfplots/table/@cell content/.add={}{&}}
%             \fi
%         }%
%     }
% }
%
\pgfplotstableread[header=true]{
road & sidew. & build. & wall & fence & pole & light & sign & veget. & terrain & sky & person & rider & car & truck & bus & train & motorc. & bicycle & mean \\
64.3 & 25.4 & 63.3 & 26.5 & 18 & 5.5 & 6.4 & 9.3 & 63.7 & 13.3 & 79.1 & 6.9 & 0.8 & 23.8 & 24.8 & 39.5 & 8.3 & 6.7 & 23.7 & 26.8 \\ % -
86.9 & 60 & 82.7 & 49 & 32.3 & 43.9 & 58.1 & 40.7 & 83.3 & 38.3 & 94.4 & 12.8 & 7.2 & 52.7 & 43.4 & 50.4 & 15.1 & 30 & 43.6 & 48.7 \\ % align
67.6 & 52 & 83.8 & 47.8 & 36.7 & 56 & 69.3 & 51.7 & 73.7 & 37 & 63.7 & 46 & 27.2 & 78.9 & 67.6 & 75.1 & 83.2 & 42.6 & 48.7 & 58.3 \\ % align, P
% 58.7 & 50.5 & 82.7 & 43.3 & 33.9 & 56.8 & 70.1 & 53.4 & 71.7 & 35.4 & 51 & 51.7 & 26.9 & 79.3 & 67.7 & 75.9 & 84 & 41.7 & 40.4 & 56.6 \\ % align, P, s
87.8 & 57.3 & 84.1 & 47.7 & 33.1 & 55.3 & 69.8 & 51.9 & 84.5 & 37.8 & 94.9 & 45.4 & 28.5 & 78.3 & 68.7 & 78.7 & 83.2 & 45 & 49.2 & 62.2 \\ % align, P, M
88.5 & 58.9 & 85.0 & 48.4 & 34.2 & 57.2 & 71.3 & 54.1 & 85.2 & 40.1 & 95.1 & 55.4 & 36.5 & 82.9 & 67.6 & 79.3 & 83.4 & 45.5 & 47.9 & 64.0 \\ % align, P, s, M
89.4 & 62.4 & 85.5 & 48.6 & 36.6 & 57.7 & 71.0 & 55.0 & 85.3 & 41.0 & 95.1 & 57.3 & 33.1 & 82.9 & 73.6 & 82.5 & 86.0 & 43.9 & 48.1 & 65.0 \\ % align, P, s, M, co
}\mytable%
%c
\def\mycolw{\textnormal{\textsc{align}} }% string which defines col width
\begin{tabular}{@{}c@{}c@{}c@{}c@{}c@{}c@{}c@{}c@{}c@{}c@{}c@{}c@{}c@{}c@{}c@{}c@{}c@{}c@{}c@{}c@{}c@{}c@{}c@{}c@{}c@{}r@{}}\toprule
& \textover[c]{\textnormal{\textsc{align}}}{\mycolw} & \textover[c]{$P_\mathcal{R}$}{\mycolw} & \textover[c]{$\mathbf{M}$}{\mycolw} & \textover[c]{$s$}{\mycolw} & \textover[c]{$\mathcal{R}$-ad}{\mycolw} & \rotatebox[origin=c]{90}{road} & \rotatebox[origin=c]{90}{sidew.} & \rotatebox[origin=c]{90}{build.} & \rotatebox[origin=c]{90}{wall} & \rotatebox[origin=c]{90}{fence} & \rotatebox[origin=c]{90}{pole} & \rotatebox[origin=c]{90}{light} & \rotatebox[origin=c]{90}{sign} & \rotatebox[origin=c]{90}{veget.} & \rotatebox[origin=c]{90}{terrain} & \rotatebox[origin=c]{90}{sky} & \rotatebox[origin=c]{90}{person} & \rotatebox[origin=c]{90}{rider} & \rotatebox[origin=c]{90}{car} & \rotatebox[origin=c]{90}{truck} & \rotatebox[origin=c]{90}{bus} & \rotatebox[origin=c]{90}{train} & \rotatebox[origin=c]{90}{motorc.} & \rotatebox[origin=c]{90}{bicycle} & \multicolumn{1}{c}{\phantom{0}\rotatebox[origin=c]{90}{\textbf{mean}}} \\
{\pgfplotstabletypeset[string type]{
\textcolor{gray}{1} \\ 
\textcolor{gray}{2} \\ 
\textcolor{gray}{3} \\ 
\textcolor{gray}{4} \\ 
\textcolor{gray}{5} \\ 
\textcolor{gray}{6} \\ 
}} &
{\pgfplotstabletypeset[string type]{
\textover[c]{\phantom{\checkmark}}{\mycolw} \\ 
\checkmark \\ 
\checkmark \\
\checkmark \\
\checkmark \\
\checkmark \\
}} &
{\pgfplotstabletypeset[string type]{
\textover[c]{\phantom{\checkmark}}{\mycolw} \\ 
\phantom{\checkmark} \\ 
\checkmark \\
\checkmark \\
\checkmark \\
\checkmark \\
}} &
{\pgfplotstabletypeset[string type]{
\textover[c]{\phantom{\checkmark}}{\mycolw} \\ 
\phantom{\checkmark} \\ 
\phantom{\checkmark} \\ 
\checkmark \\
\checkmark \\
\checkmark \\
}} &
{\pgfplotstabletypeset[string type]{
\textover[c]{\phantom{\checkmark}}{\mycolw} \\ 
\phantom{\checkmark} \\ 
\phantom{\checkmark} \\ 
\phantom{\checkmark} \\ 
\checkmark \\
\checkmark \\
}} &
{\pgfplotstabletypeset[string type]{
\textover[c]{\phantom{\checkmark}}{\mycolw} \\ 
\phantom{\checkmark} \\ 
\phantom{\checkmark} \\ 
\phantom{\checkmark} \\ 
\phantom{\checkmark} \\ 
\checkmark \\
}} &
{\pgfplotstabletypeset[
    columns={road},
    color cells={min=\roadmin,max=\roadmax},
    fixed zerofill,
    precision=1,
]\mytable} &
{\pgfplotstabletypeset[
    columns={sidew.},
    color cells={min=\sidewmin,max=\sidewmax},
    fixed zerofill,
    precision=1,
]\mytable} &
{\pgfplotstabletypeset[
    columns={build.},
    color cells={min=\buildmin,max=\buildmax},
    fixed zerofill,
    precision=1,
]\mytable} &
{\pgfplotstabletypeset[
    columns={wall},
    color cells={min=\wallmin,max=\wallmax},
    fixed zerofill,
    precision=1,
]\mytable} &
{\pgfplotstabletypeset[
    columns={fence},
    color cells={min=\fencemin,max=\fencemax},
    fixed zerofill,
    precision=1,
]\mytable} &
{\pgfplotstabletypeset[
    columns={pole},
    fixed zerofill,
    precision=1,
    % dec sep align,
    % string type in dec sep align,
    color cells={min=\polemin,max=\polemax},
]\mytable} &
{\pgfplotstabletypeset[
    columns={light},
    color cells={min=\lightmin,max=\lightmax},
    fixed zerofill,
    precision=1,
]\mytable} &
{\pgfplotstabletypeset[
    columns={sign},
    color cells={min=\signmin,max=\signmax},
    fixed zerofill,
    precision=1,
]\mytable} &
{\pgfplotstabletypeset[
    columns={veget.},
    color cells={min=\vegetmin,max=\vegetmax},
    fixed zerofill,
    precision=1,
]\mytable} &
{\pgfplotstabletypeset[
    columns={terrain},
    color cells={min=\terrainmin,max=\terrainmax},
    fixed zerofill,
    precision=1,
]\mytable} &
{\pgfplotstabletypeset[
    columns={sky},
    color cells={min=\skymin,max=\skymax},
    fixed zerofill,
    precision=1,
]\mytable} &
{\pgfplotstabletypeset[
    columns={person},
    color cells={min=\personmin,max=\personmax},
    fixed zerofill,
    precision=1,
]\mytable} &
{\pgfplotstabletypeset[
    columns={rider},
    fixed,
    fixed zerofill,
    precision=1,
    % dec sep align={c},
    % string type in dec sep align,
    color cells={min=\ridermin,max=\ridermax},
]\mytable} &
{\pgfplotstabletypeset[
    columns={car},
    color cells={min=\carmin,max=\carmax},
    fixed zerofill,
    precision=1,
]\mytable} &
{\pgfplotstabletypeset[
    columns={truck},
    color cells={min=\truckmin,max=\truckmax},
    fixed zerofill,
    precision=1,
]\mytable} &
{\pgfplotstabletypeset[
    columns={bus},
    color cells={min=\busmin,max=\busmax},
    fixed zerofill,
    precision=1,
]\mytable} &
{\pgfplotstabletypeset[
    columns={train},
    color cells={min=\trainmin,max=\trainmax},
    fixed zerofill,
    precision=1,
]\mytable} &
{\pgfplotstabletypeset[
    columns={motorc.},
    color cells={min=\motorcmin,max=\motorcmax},
    fixed zerofill,
    precision=1,
]\mytable} &
{\pgfplotstabletypeset[
    columns={bicycle},
    color cells={min=\bicyclemin,max=\bicyclemax},
    fixed zerofill,
    precision=1,
]\mytable} &
{\pgfplotstabletypeset[
    columns={mean},
    color cells={min=\meanmin,max=\meanmax},
    fixed zerofill,
    precision=1,
    columns/0/.style={column type=r},%
]\mytable}%
\end{tabular}
    }
\label{tab:ablations}
\vspace{-3mm}
\end{table*}

\begin{table}
\caption{Hyperparameter study of the mean-entropy exponent $\gamma$ on the ACDC validation set.}%
\smallskip%
\centering%
\resizebox{\linewidth}{!}{%
\ra{1.3}%
\setlength{\tabcolsep}{14pt}%
\begin{tabular}{@{}lcccccr@{}}\toprule
&& \multicolumn{5}{c}{mIoU\,$\uparrow$} \\
\cmidrule{3-7}
&& \(\gamma=1\) & \( \gamma=\frac{1}{2} \) & \( \gamma=\frac{1}{4} \) & \( \gamma=\frac{1}{8} \) & \( \gamma=\frac{1}{16} \) \\
\cmidrule{1-7} 
Refign-DAFormer && 59.2 & 61.8 & \textbf{65.0} & 64.3 & 63.5 \\
\bottomrule
\end{tabular}%
}
\label{tab:power_ablation}
\end{table}

\begin{figure}
    \begin{minipage}[c]{0.49\linewidth}%
        \adjustbox{max width=\linewidth}{%
            \input{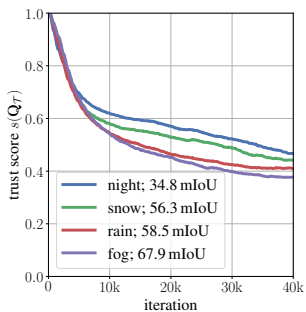}%
        }%
    \end{minipage}\hfill%
    \begin{minipage}[c]{0.49\linewidth}%
    \caption{Trust score $s(\mathbf{Q}_{\mathcal{T}})$ during training for images of the different conditions in ACDC. On average, difficult conditions (night, snow\textemdash lower mIoU) exhibit a higher $s$, meaning their target predictions undergo more intensive correction by reference predictions. Shown are mIoU validation scores of the DAFormer~\cite{hoyer2021daformer} baseline.}%
    \label{fig:trust_score}
    \end{minipage}
\vspace{-2mm}
\end{figure}

\begin{figure}
    \begin{minipage}[c]{0.49\linewidth}%
        \adjustbox{max width=\linewidth}{%
            \input{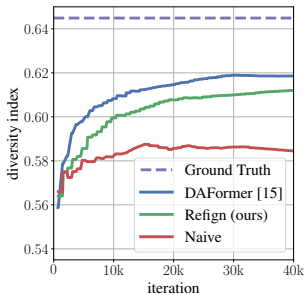}%
        }%
    \end{minipage}\hfill%
    \begin{minipage}[c]{0.49\linewidth}%
    \caption{Prediction diversity on the ACDC validation set during training. We use normalized entropy as a diversity index. Refign preserves higher prediction diversity compared to a naive refinement scheme, which consists of simple averaging ($\alpha=0.5$ in \eqref{eq:convex_comb}).}
    \label{fig:diversity}
    \end{minipage}
    \vspace{-5mm}
\end{figure}

\PAR{ACDC.}
We present comparisons to several state-of-the-art methods on the ACDC test set in Table~\ref{tab:acdc_sota}.
Applying Refign on top of DAFormer~\cite{hoyer2021daformer} results in a mIoU of 65.5\%, setting the new state of the art in domain adaptation from Cityscapes to ACDC.
Refign boosts the performance of DAFormer by a substantial 10.1\%.
Besides static classes, we observe substantial improvement for dynamic classes as well, owing to our adaptive refinement method.
Among DeepLabv2-based methods, our method Refign-DACS is the second best after DANIA~\cite{wu2021one}.
Note that Refign boosts the mIoU of DACS~\cite{tranheden2021dacs} by 6.8\%.

We present qualitative comparisons with FDA, DANIA, and DAFormer in Fig.~\ref{fig:qualitative_seg}.
Our Refign-DAFormer consistently produces more accurate segmentation maps than the other methods.
For instance, Refign corrects typical misclassifications of DAFormer, \eg \emph{sky} as \emph{road}.

\PAR{Dark Zurich.}
In Table~\ref{tab:night_generalization}, we benchmark our method on Dark Zurich-test.
Following previous works, the trained Dark Zurich models are also tested for generalization on Nighttime Driving~\cite{dai2018dark} and BDD100k-night~\cite{yu2020bdd100k,sakaridis2020map}.
Refign markedly improves over the DACS and DAFormer baselines, both on Dark Zurich-test and the two unseen domains.
Notably, Refign-DAFormer achieves a mIoU of 56.2\% on Dark Zurich, setting the new state of the art.

\subsection{Semi-Supervised Domain Adaptation Results}
\label{sec:exp:robotcar_cmu}

Table~\ref{tab:robotcar_cmu} lists semi-supervised domain adaptation results on the RobotCar and CMU
Correspondence datasets.
We compare to DAFormer~\cite{hoyer2021daformer} and the three PSPNet-based~\cite{zhao2017pyramid} models proposed in~\cite{larsson2019cross}.
The models in \cite{larsson2019cross} rely on sparse 2D matches obtained via a multi-stage pipeline involving camera pose estimation, 3D point cloud generation and matching, match pruning, and 2D reprojection.
In contrast, our alignment module directly establishes correspondences through a dense matching network.
Our models achieve the best score on both datasets, demonstrating the generality of our approach.

\begin{table*}
\caption{Comparison to the state of the art in geometric matching. All methods are trained on MegaDepth and evaluated on MegaDepth, RobotCar, and CMU. ``w/o SfM'': trained without sparse structure from motion matches, ``UA'': uncertainty-aware matching network.}%
\smallskip%
\centering%
\resizebox{\linewidth}{!}{%
\ra{1.3}%
\setlength{\tabcolsep}{9pt}
\begin{tabular}{@{}lccccccccccccccccccr@{}}\toprule
\multirow{2}{*}{Method} && \multirow{2}{*}{w/o SfM} && \multirow{2}{*}{UA} && \multicolumn{4}{c}{MegaDepth~\cite{li2018megadepth}} && \multicolumn{4}{c}{RobotCar~\cite{maddern20171,larsson2019cross}} && \multicolumn{4}{c}{CMU~\cite{badino2011visual,larsson2019cross}} \\
\cmidrule{7-10} \cmidrule{12-15} \cmidrule{17-20} &&&&&& PCK-1\,$\uparrow$ & PCK-5\,$\uparrow$ & PCK-10\,$\uparrow$ & AUSE\,$\downarrow$ && PCK-1\,$\uparrow$ & PCK-5\,$\uparrow$ & PCK-10\,$\uparrow$ & AUSE\,$\downarrow$ && PCK-1\,$\uparrow$ & PCK-5\,$\uparrow$ & PCK-10\,$\uparrow$ & AUSE\,$\downarrow$ \\ \midrule
% SIFT-Flow~\cite{liu2010sift} &&&&&& \phantom{0}1.12 & \phantom{0}8.13 & 16.45 & --- && & & & ---  \\
% NCNet~\cite{rocco2018neighbourhood} &&&&&& \phantom{0}0.81 & \phantom{0}7.13 & 16.93 & --- && & & & ---  \\
DGC+M-Net~\cite{melekhov2019dgc} && \checkmark && \checkmark && \phantom{0}4.10 & 33.60 & 49.39 & 0.320 && \phantom{0}1.11 & 19.12 & 38.92 & 0.241 && \phantom{0}1.99 & 27.15 & 51.99 & 0.320  \\
% RANSAC-Flow~\cite{shen2020ransac} &&&&&& \phantom{0}2.10 & 16.07 & 31.66 & --- && & & & ---  \\
GLU-Net~\cite{truong2020glu} && \checkmark &&&& 29.46 & 55.96 & 62.39 & - && \phantom{0}2.21 & 33.72 & 55.28 & - && 21.18 & 80.95 & 91.44 & -\phantom{0\,}  \\
% GLU-Net\textsubscript{dyn}~\cite{truong2020glu,truong2020gocor} &&&&&& 21.59 & 61.90 & 69.81 & 0.253 && \phantom{0}2.31 & 33.91 & 55.69 & 0.218 && 19.75 & 84.36 & 95.20 & 0.279   \\ 
WarpC~\cite{truong2021warp} && \checkmark &&&& 50.86 & 78.76 & 83.00 & - && \phantom{0}2.51 & 35.93 & 57.45 & - && 24.74 & 86.10 & 95.65 & -\phantom{0\,} \\
% PDC-Net~\cite{truong2021learning} &&&& \checkmark &&  &  &  &  && \textbf{\phantom{0}2.54} & 36.37 & 57.69 & 0.199 && \textbf{28.04} & \textbf{86.94} & 94.37 & 0.294  \\
PDC-Net+~\cite{truong2021pdc} &&&& \checkmark && \textbf{72.42} & \textbf{88.10} & \textbf{89.31} & 0.293 && \phantom{0}2.57 & 36.71 & \textbf{58.44} & 0.186 && \textbf{27.84} & 85.21 & 92.57 & \textbf{0.268}  \\
\midrule
UAWarpC (ours) && \checkmark && \checkmark && 53.04 & 78.52 & 81.92 & \textbf{0.217} && \textbf{\phantom{0}2.59} & \textbf{36.79} & 58.10 & \textbf{0.155} && 27.44 & \textbf{88.17} & \textbf{95.94} & 0.270
\\
\bottomrule
\end{tabular}%
}
\label{tab:densecorr}
\vspace{-3mm}
\end{table*}

\begin{figure*}
    \adjustbox{max width=\textwidth}{%
        \input{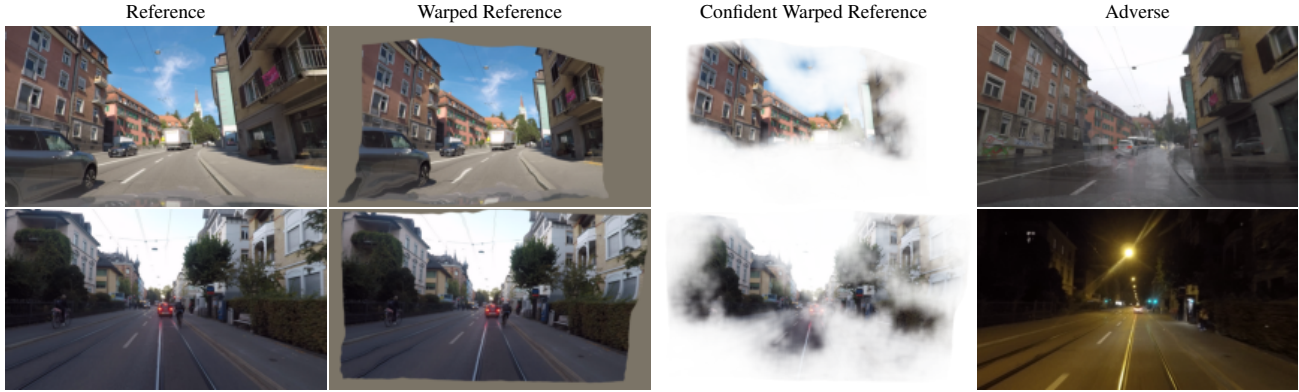}%
    }%
    \caption{Visualizations of warped reference images and the corresponding confidence maps (white $\leftrightarrow$ low confidence) from ACDC.}
    \label{fig:qualitative_warp}
    \vspace{-5mm}
\end{figure*}

\subsection{Ablation Study and Further Analysis}
\label{sec:exp:ablations}

We perform an extensive analysis of our method on the ACDC validation set.
To obtain more reliable performance estimates, all experiments in this section are repeated three times and we report the average performance.

Table~\ref{tab:ablations} shows the ablation study of different components of our refinement scheme \eqref{eq:epsilon_2}.
The first row lists a naive refinement scheme, where $\alpha=0.5$ and no alignment is performed.
Adding alignment (row 2; +21.9\%\,mIoU) improves performance substantially across all classes.
Further adding $P_\mathcal{R}$ (row 3; +9.6\%\,mIoU) sharply improves performance for dynamic classes and small static classes, but some large static classes are insufficiently mixed and their performance decreases due to low warping confidence within non-textured regions (Sec.~\ref{sec:method:refinement}).
Performance for large static classes is fully recovered by incorporating aggressive large static class mixing via $\mathbf{M}$, without sacrificing performance on small static classes or dynamic classes (row 4; +3.9\%\,mIoU).
Introducing confidence-adaptive mixing with trust score $s$ increases performance further for 17 of the 19 classes (row 5; +1.8\%\,mIoU).
We arrive at the final configuration by adding concurrent adaptation to $\mathcal{R}$\textemdash as detailed in Lines~\ref{alglst:adr_start}-\ref{alglst:adr_end} of Alg.~\ref{alg:refign}\textemdash which adds another +1.0\%\,mIoU in performance.

Table~\ref{tab:power_ablation} shows the sensitivity of our method to different values of its only hyperparameter $\gamma$ (mean-entropy exponent, see \eqref{eq:gamma_eq}).
The method is fairly robust to decreasing $\gamma$.
In Fig.~\ref{fig:trust_score}, we compare values of trust score $s$ during training across different conditions. We observe that $s$ remains larger for images of more challenging conditions (night, snow).
Thus, the refinement scheme automatically relies more heavily on the reference predictions for these conditions.
Finally, we plot in Fig.~\ref{fig:diversity} the diversity of validation set predictions during training, measured by the normalized entropy of the predicted class histogram.
Compared to a naive mixing scheme consisting of simple averaging, Refign preserves considerably higher prediction diversity.

\subsection{Geometric Matching Results}
\label{sec:exp:matching}

In Table~\ref{tab:densecorr} we compare our alignment strategy UAWarpC to state-of-the-art geometric matching methods.
All reported methods are trained on MegaDepth.
To estimate their suitability for deployment as an alignment module in Refign, we report generalization performance on two road datasets, RobotCar and CMU.
UAWarpC improves the generalization accuracy compared to the WarpC~\cite{truong2021warp} baseline, demonstrating that our uncertainty modeling increases the robustness of the flow estimator.
In terms of uncertainty estimation, our method matches or even outperforms the recent probabilistic PDC-Net+~\cite{truong2021pdc} in AUSE.

Finally, we show qualitative warp results in Fig.~\ref{fig:qualitative_warp}.
Notably, the alignment module successfully manages to identify dynamic objects and assigns them a low confidence.

%%%%%%%%% CONCLUSION
\section{Conclusion}

We present Refign, a generic add-on to self-training-based UDA methods that leverages an additional \emph{reference} image for each target-domain image.
Refign consists of two steps:
(1) uncertainty-aware alignment of the reference prediction with the target prediction, and
(2) adaptive refinement of the target predictions with the aligned reference predictions.
To enable step (1), we propose UAWarpC, a probabilistic extension to the matching method WarpC~\cite{truong2021warp}.
UAWarpC reaches state-of-the-art performance in both flow accuracy and uncertainty estimation.
Step (2) consists of a non-parametric label correction scheme.
We apply Refign to two existing UDA methods\textemdash DACS~\cite{tranheden2021dacs} and DAFormer~\cite{hoyer2021daformer}\textemdash and report state-of-the-art normal-to-adverse domain adaptation results for semantic segmentation on multiple driving scene datasets.

\PAR{Acknowledgment.}  
This work was supported by the ETH Future Computing Laboratory (EFCL), financed by a donation from Huawei Technologies.

{\small
\bibliographystyle{ieee_fullname}
\bibliography{refs}

\begin{thebibliography}{10}\itemsep=-1pt

\bibitem{arpit2017closer}
Devansh Arpit, Stanislaw Jastrzebski, Nicolas Ballas, David Krueger, Emmanuel
  Bengio, Maxinder~S Kanwal, Tegan Maharaj, Asja Fischer, Aaron Courville,
  Yoshua Bengio, et~al.
\newblock A closer look at memorization in deep networks.
\newblock In {\em ICML}, 2017.

\bibitem{badino2011visual}
Hern{\'a}n Badino, Daniel Huber, and Takeo Kanade.
\newblock Visual topometric localization.
\newblock In {\em IEEE Intelligent vehicles symposium (IV)}, 2011.

\bibitem{bengio2009curriculum}
Yoshua Bengio, J{\'e}r{\^o}me Louradour, Ronan Collobert, and Jason Weston.
\newblock Curriculum learning.
\newblock In {\em ICML}, 2009.

\bibitem{burnett2022boreas}
Keenan Burnett, David~J Yoon, Yuchen Wu, Andrew~Zou Li, Haowei Zhang, Shichen
  Lu, Jingxing Qian, Wei-Kang Tseng, Andrew Lambert, Keith~YK Leung, et~al.
\newblock Boreas: A multi-season autonomous driving dataset, 2022.

\bibitem{chen2017deeplab}
Liang-Chieh Chen, George Papandreou, Iasonas Kokkinos, Kevin Murphy, and Alan~L
  Yuille.
\newblock Deeplab: Semantic image segmentation with deep convolutional nets,
  atrous convolution, and fully connected crfs.
\newblock {\em IEEE TPAMI}, 40(4):834--848, 2017.

\bibitem{cordts2016cityscapes}
Marius Cordts, Mohamed Omran, Sebastian Ramos, Timo Rehfeld, Markus Enzweiler,
  Rodrigo Benenson, Uwe Franke, Stefan Roth, and Bernt Schiele.
\newblock The cityscapes dataset for semantic urban scene understanding.
\newblock In {\em CVPR}, 2016.

\bibitem{dai2020curriculum}
Dengxin Dai, Christos Sakaridis, Simon Hecker, and Luc Van~Gool.
\newblock Curriculum model adaptation with synthetic and real data for semantic
  foggy scene understanding.
\newblock {\em IJCV}, 128(5):1182--1204, 2020.

\bibitem{dai2018dark}
Dengxin Dai and Luc Van~Gool.
\newblock Dark model adaptation: Semantic image segmentation from daytime to
  nighttime.
\newblock In {\em International Conference on Intelligent Transportation
  Systems (ITSC)}, 2018.

\bibitem{gao2022cross}
Huan Gao, Jichang Guo, Guoli Wang, and Qian Zhang.
\newblock Cross-domain correlation distillation for unsupervised domain
  adaptation in nighttime semantic segmentation.
\newblock In {\em CVPR}, 2022.

\bibitem{gong2012geodesic}
Boqing Gong, Yuan Shi, Fei Sha, and Kristen Grauman.
\newblock Geodesic flow kernel for unsupervised domain adaptation.
\newblock In {\em CVPR}, 2012.

\bibitem{gopalan2011domain}
Raghuraman Gopalan, Ruonan Li, and Rama Chellappa.
\newblock Domain adaptation for object recognition: An unsupervised approach.
\newblock In {\em ICCV}, 2011.

\bibitem{gou2021knowledge}
Jianping Gou, Baosheng Yu, Stephen~J Maybank, and Dacheng Tao.
\newblock Knowledge distillation: A survey.
\newblock {\em IJCV}, 129(6):1789--1819, 2021.

\bibitem{cyCADA}
Judy Hoffman, Eric Tzeng, Taesung Park, Jun-Yan Zhu, Phillip Isola, Kate
  Saenko, Alexei Efros, and Trevor Darrell.
\newblock {CyCADA}: Cycle-consistent adversarial domain adaptation.
\newblock In {\em ICML}, 2018.

\bibitem{FCNs:in:the:wild}
Judy Hoffman, Dequan Wang, Fisher Yu, and Trevor Darrell.
\newblock Fcns in the wild: Pixel-level adversarial and constraint-based
  adaptation, 2016.

\bibitem{hoyer2021daformer}
Lukas Hoyer, Dengxin Dai, and Luc Van~Gool.
\newblock Daformer: Improving network architectures and training strategies for
  domain-adaptive semantic segmentation.
\newblock In {\em CVPR}, 2022.

\bibitem{hoyer2022hrda}
Lukas Hoyer, Dengxin Dai, and Luc Van~Gool.
\newblock Hrda: Context-aware high-resolution domain-adaptive semantic
  segmentation, 2022.

\bibitem{huber1992robust}
Peter~J Huber.
\newblock Robust estimation of a location parameter.
\newblock In {\em Breakthroughs in statistics}, pages 492--518. Springer, 1992.

\bibitem{jiang2021cotr}
Wei Jiang, Eduard Trulls, Jan Hosang, Andrea Tagliasacchi, and Kwang~Moo Yi.
\newblock {COTR:} correspondence transformer for matching across images.
\newblock In {\em ICCV}, 2021.

\bibitem{kamann2020benchmarking}
Christoph Kamann and Carsten Rother.
\newblock Benchmarking the robustness of semantic segmentation models.
\newblock In {\em CVPR}, 2020.

\bibitem{textda:adaptation}
Myeongjin Kim and Hyeran Byun.
\newblock Learning texture invariant representation for domain adaptation of
  semantic segmentation.
\newblock In {\em CVPR}, 2020.

\bibitem{kingma2014adam}
Diederik~P Kingma and Jimmy Ba.
\newblock Adam: A method for stochastic optimization.
\newblock In {\em ICLR}, 2015.

\bibitem{larsson2019cross}
Mans Larsson, Erik Stenborg, Lars Hammarstrand, Marc Pollefeys, Torsten
  Sattler, and Fredrik Kahl.
\newblock A cross-season correspondence dataset for robust semantic
  segmentation.
\newblock In {\em CVPR}, 2019.

\bibitem{lee2022fifo}
Sohyun Lee, Taeyoung Son, and Suha Kwak.
\newblock {FIFO}: Learning fog-invariant features for foggy scene segmentation.
\newblock In {\em CVPR}, 2022.

\bibitem{li2020dual}
Xinghui Li, Kai Han, Shuda Li, and Victor Prisacariu.
\newblock Dual-resolution correspondence networks.
\newblock In {\em NeurIPS}, 2020.

\bibitem{li2019bidirectional}
Yunsheng Li, Lu Yuan, and Nuno Vasconcelos.
\newblock Bidirectional learning for domain adaptation of semantic
  segmentation.
\newblock In {\em CVPR}, 2019.

\bibitem{li2018megadepth}
Zhengqi Li and Noah Snavely.
\newblock Megadepth: Learning single-view depth prediction from internet
  photos.
\newblock In {\em CVPR}, 2018.

\bibitem{liao2022unsupervised}
Liang Liao, Wenyi Chen, Jing Xiao, Zheng Wang, Chia-Wen Lin, and Shin’ichi
  Satoh.
\newblock Unsupervised foggy scene understanding via self spatial-temporal
  label diffusion.
\newblock {\em IEEE Transactions on Image Processing}, 31:3525--3540, 2022.

\bibitem{loshchilov2017decoupled}
Ilya Loshchilov and Frank Hutter.
\newblock Decoupled weight decay regularization.
\newblock In {\em ICLR}, 2019.

\bibitem{luo2019taking}
Yawei Luo, Liang Zheng, Tao Guan, Junqing Yu, and Yi Yang.
\newblock Taking a closer look at domain shift: Category-level adversaries for
  semantics consistent domain adaptation.
\newblock In {\em CVPR}, 2019.

\bibitem{ma2021both}
Xianzheng Ma, Zhixiang Wang, Yacheng Zhan, Yinqiang Zheng, Zheng Wang, Dengxin
  Dai, and Chia-Wen Lin.
\newblock Both style and fog matter: Cumulative domain adaptation for semantic
  foggy scene understanding.
\newblock In {\em CVPR}, 2022.

\bibitem{maddern20171}
Will Maddern, Geoffrey Pascoe, Chris Linegar, and Paul Newman.
\newblock 1 year, 1000 km: The oxford robotcar dataset.
\newblock {\em The International Journal of Robotics Research}, 36(1):3--15,
  2017.

\bibitem{melekhov2019dgc}
Iaroslav Melekhov, Aleksei Tiulpin, Torsten Sattler, Marc Pollefeys, Esa Rahtu,
  and Juho Kannala.
\newblock {DGC-Net}: Dense geometric correspondence network.
\newblock In {\em WACV}, 2019.

\bibitem{nix1994estimating}
David~A Nix and Andreas~S Weigend.
\newblock Estimating the mean and variance of the target probability
  distribution.
\newblock In {\em Proceedings of 1994 ieee international conference on neural
  networks (ICNN'94)}, volume~1, pages 55--60, 1994.

\bibitem{porav2019don}
Horia Porav, Tom Bruls, and Paul Newman.
\newblock Don’t worry about the weather: Unsupervised condition-dependent
  domain adaptation.
\newblock In {\em IEEE Intelligent Transportation Systems Conference (ITSC)},
  2019.

\bibitem{reed2014training}
Scott Reed, Honglak Lee, Dragomir Anguelov, Christian Szegedy, Dumitru Erhan,
  and Andrew Rabinovich.
\newblock Training deep neural networks on noisy labels with bootstrapping.
\newblock In {\em ICLR}, 2015.

\bibitem{rocco2018neighbourhood}
Ignacio Rocco, Mircea Cimpoi, Relja Arandjelovi{\'c}, Akihiko Torii, Tomas
  Pajdla, and Josef Sivic.
\newblock Neighbourhood consensus networks.
\newblock In {\em NeurIPS}, 2018.

\bibitem{romera2019bridging}
Eduardo Romera, Luis~M. Bergasa, Kailun Yang, Jose~M. Alvarez, and Rafael
  Barea.
\newblock Bridging the day and night domain gap for semantic segmentation.
\newblock In {\em IEEE Intelligent Vehicles Symposium (IV)}, 2019.

\bibitem{sakaridis2019guided}
Christos Sakaridis, Dengxin Dai, and Luc~Van Gool.
\newblock Guided curriculum model adaptation and uncertainty-aware evaluation
  for semantic nighttime image segmentation.
\newblock In {\em ICCV}, 2019.

\bibitem{SynRealDataFogECCV18}
Christos Sakaridis, Dengxin Dai, Simon Hecker, and Luc Van~Gool.
\newblock Model adaptation with synthetic and real data for semantic dense
  foggy scene understanding.
\newblock In {\em ECCV}, 2018.

\bibitem{sakaridis2018semantic}
Christos Sakaridis, Dengxin Dai, and Luc Van~Gool.
\newblock Semantic foggy scene understanding with synthetic data.
\newblock {\em IJCV}, 126(9):973--992, 2018.

\bibitem{sakaridis2021acdc}
Christos Sakaridis, Dengxin Dai, and Luc Van~Gool.
\newblock {ACDC}: The {Adverse} {Conditions} {Dataset} with {Correspondences}
  for semantic driving scene understanding.
\newblock In {\em ICCV}, 2021.

\bibitem{sakaridis2020map}
Christos Sakaridis, Dengxin Dai, and Luc Van~Gool.
\newblock Map-guided curriculum domain adaptation and uncertainty-aware
  evaluation for semantic nighttime image segmentation.
\newblock {\em TPAMI}, 44(6):3139--3153, 2022.

\bibitem{synthetic:semantic:segmentation}
Swami Sankaranarayanan, Yogesh Balaji, Arpit Jain, Ser Nam~Lim, and Rama
  Chellappa.
\newblock Learning from synthetic data: Addressing domain shift for semantic
  segmentation.
\newblock In {\em CVPR}, 2018.

\bibitem{schonberger2016structure}
Johannes~L Schonberger and Jan-Michael Frahm.
\newblock Structure-from-motion revisited.
\newblock In {\em CVPR}, 2016.

\bibitem{shen2020ransac}
Xi Shen, Fran{\c{c}}ois Darmon, Alexei~A Efros, and Mathieu Aubry.
\newblock Ransac-flow: generic two-stage image alignment.
\newblock In {\em ECCV}, 2020.

\bibitem{simonyan2014very}
Karen Simonyan and Andrew Zisserman.
\newblock Very deep convolutional networks for large-scale image recognition.
\newblock In {\em ICLR}, 2015.

\bibitem{song2019selfie}
Hwanjun Song, Minseok Kim, and Jae-Gil Lee.
\newblock Selfie: Refurbishing unclean samples for robust deep learning.
\newblock In {\em ICML}, 2019.

\bibitem{sun2019see}
Lei Sun, Kaiwei Wang, Kailun Yang, and Kaite Xiang.
\newblock See clearer at night: towards robust nighttime semantic segmentation
  through day-night image conversion.
\newblock In {\em Artificial Intelligence and Machine Learning in Defense
  Applications}, 2019.

\bibitem{tanaka2018joint}
Daiki Tanaka, Daiki Ikami, Toshihiko Yamasaki, and Kiyoharu Aizawa.
\newblock Joint optimization framework for learning with noisy labels.
\newblock In {\em CVPR}, 2018.

\bibitem{tarvainen2017mean}
Antti Tarvainen and Harri Valpola.
\newblock Mean teachers are better role models: Weight-averaged consistency
  targets improve semi-supervised deep learning results.
\newblock In {\em NeurIPS}, 2017.

\bibitem{tranheden2021dacs}
Wilhelm Tranheden, Viktor Olsson, Juliano Pinto, and Lennart Svensson.
\newblock Dacs: Domain adaptation via cross-domain mixed sampling.
\newblock In {\em WACV}, 2021.

\bibitem{tremblay2021rain}
Maxime Tremblay, Shirsendu~Sukanta Halder, Raoul De~Charette, and
  Jean-Fran{\c{c}}ois Lalonde.
\newblock Rain rendering for evaluating and improving robustness to bad
  weather.
\newblock {\em IJCV}, 129(2):341--360, 2021.

\bibitem{truong2020glu}
Prune Truong, Martin Danelljan, and Radu Timofte.
\newblock Glu-net: Global-local universal network for dense flow and
  correspondences.
\newblock In {\em CVPR}, 2020.

\bibitem{truong2021pdc}
Prune Truong, Martin Danelljan, Radu Timofte, and Luc Van~Gool.
\newblock Pdc-net+: Enhanced probabilistic dense correspondence network, 2021.

\bibitem{truong2021warp}
Prune Truong, Martin Danelljan, Fisher Yu, and Luc Van~Gool.
\newblock Warp consistency for unsupervised learning of dense correspondences.
\newblock In {\em ICCV}, 2021.

\bibitem{truong2022probabilistic}
Prune Truong, Martin Danelljan, Fisher Yu, and Luc Van~Gool.
\newblock Probabilistic warp consistency for weakly-supervised semantic
  correspondences.
\newblock In {\em CVPR}, 2022.

\bibitem{tsai2018learning}
Yi-Hsuan Tsai, Wei-Chih Hung, Samuel Schulter, Kihyuk Sohn, Ming-Hsuan Yang,
  and Manmohan Chandraker.
\newblock Learning to adapt structured output space for semantic segmentation.
\newblock In {\em CVPR}, 2018.

\bibitem{patch:align:adaptation}
Yi-Hsuan Tsai, Kihyuk Sohn, Samuel Schulter, and Manmohan Chandraker.
\newblock Domain adaptation for structured output via discriminative patch
  representations.
\newblock In {\em ICCV}, 2019.

\bibitem{vu2019advent}
Tuan-Hung Vu, Himalaya Jain, Maxime Bucher, Matthieu Cord, and Patrick
  P{\'e}rez.
\newblock Advent: Adversarial entropy minimization for domain adaptation in
  semantic segmentation.
\newblock In {\em CVPR}, 2019.

\bibitem{wang2021proselflc}
Xinshao Wang, Yang Hua, Elyor Kodirov, David~A Clifton, and Neil~M Robertson.
\newblock Proselflc: Progressive self label correction for training robust deep
  neural networks.
\newblock In {\em CVPR}, 2021.

\bibitem{wang2020differential}
Zhonghao Wang, Mo Yu, Yunchao Wei, Rogerio Feris, Jinjun Xiong, Wen-mei Hwu,
  Thomas~S Huang, and Honghui Shi.
\newblock Differential treatment for stuff and things: A simple unsupervised
  domain adaptation method for semantic segmentation.
\newblock In {\em CVPR}, 2020.

\bibitem{wei2020theoretical}
Colin Wei, Kendrick Shen, Yining Chen, and Tengyu Ma.
\newblock Theoretical analysis of self-training with deep networks on unlabeled
  data.
\newblock In {\em ICLR}, 2021.

\bibitem{wu2021dannet}
Xinyi Wu, Zhenyao Wu, Hao Guo, Lili Ju, and Song Wang.
\newblock {DANNet}: A one-stage domain adaptation network for unsupervised
  nighttime semantic segmentation.
\newblock In {\em CVPR}, 2021.

\bibitem{wu2021one}
Xinyi Wu, Zhenyao Wu, Lili Ju, and Song Wang.
\newblock A one-stage domain adaptation network with image alignment for
  unsupervised nighttime semantic segmentation.
\newblock {\em IEEE TPAMI}, (01):1--1, 2021.

\bibitem{incremental:adversarial:DA:18}
Markus Wulfmeier, Alex Bewley, and Ingmar Posner.
\newblock Incremental adversarial domain adaptation for continually changing
  environments.
\newblock In {\em ICRA}, 2018.

\bibitem{xie2022sepico}
Binhui Xie, Shuang Li, Mingjia Li, Chi~Harold Liu, Gao Huang, and Guoren Wang.
\newblock Sepico: Semantic-guided pixel contrast for domain adaptive semantic
  segmentation, 2022.

\bibitem{xie2021segformer}
Enze Xie, Wenhai Wang, Zhiding Yu, Anima Anandkumar, Jose~M Alvarez, and Ping
  Luo.
\newblock Segformer: Simple and efficient design for semantic segmentation with
  transformers.
\newblock In {\em NeurIPS}, 2021.

\bibitem{xu2021cdada}
Qi Xu, Yinan Ma, Jing Wu, Chengnian Long, and Xiaolin Huang.
\newblock Cdada: A curriculum domain adaptation for nighttime semantic
  segmentation.
\newblock In {\em ICCV}, 2021.

\bibitem{yan2021deep}
Xiaoqiang Yan, Shizhe Hu, Yiqiao Mao, Yangdong Ye, and Hui Yu.
\newblock Deep multi-view learning methods: a review.
\newblock {\em Neurocomputing}, 448:106--129, 2021.

\bibitem{yang2020fda}
Yanchao Yang and Stefano Soatto.
\newblock Fda: Fourier domain adaptation for semantic segmentation.
\newblock In {\em CVPR}, 2020.

\bibitem{yarowsky1995unsupervised}
David Yarowsky.
\newblock Unsupervised word sense disambiguation rivaling supervised methods.
\newblock In {\em 33rd annual meeting of the association for computational
  linguistics}, 1995.

\bibitem{yu2020bdd100k}
Fisher Yu, Haofeng Chen, Xin Wang, Wenqi Xian, Yingying Chen, Fangchen Liu,
  Vashisht Madhavan, and Trevor Darrell.
\newblock Bdd100k: A diverse driving dataset for heterogeneous multitask
  learning.
\newblock In {\em CVPR}, 2020.

\bibitem{yuan2020revisiting}
Li Yuan, Francis~EH Tay, Guilin Li, Tao Wang, and Jiashi Feng.
\newblock Revisiting knowledge distillation via label smoothing regularization.
\newblock In {\em CVPR}, 2020.

\bibitem{zhang2021understanding}
Chiyuan Zhang, Samy Bengio, Moritz Hardt, Benjamin Recht, and Oriol Vinyals.
\newblock Understanding deep learning (still) requires rethinking
  generalization.
\newblock {\em Communications of the ACM}, 64(3):107--115, 2021.

\bibitem{zhang2021prototypical}
Pan Zhang, Bo Zhang, Ting Zhang, Dong Chen, Yong Wang, and Fang Wen.
\newblock Prototypical pseudo label denoising and target structure learning for
  domain adaptive semantic segmentation.
\newblock In {\em CVPR}, 2021.

\bibitem{FCNs:adaptation}
Yiheng Zhang, Zhaofan Qiu, Ting Yao, Dong Liu, and Tao Mei.
\newblock Fully convolutional adaptation networks for semantic segmentation.
\newblock In {\em CVPR}, 2018.

\bibitem{zhao2017pyramid}
Hengshuang Zhao, Jianping Shi, Xiaojuan Qi, Xiaogang Wang, and Jiaya Jia.
\newblock Pyramid scene parsing network.
\newblock In {\em CVPR}, 2017.

\bibitem{zheng2021rectifying}
Zhedong Zheng and Yi Yang.
\newblock Rectifying pseudo label learning via uncertainty estimation for
  domain adaptive semantic segmentation.
\newblock {\em IJCV}, 129(4):1106--1120, 2021.

\bibitem{zheng2019unsupervised}
Zhedong Zheng and Yi Yang.
\newblock Unsupervised scene adaptation with memory regularization in vivo.
\newblock In {\em International Joint Conference on Artificial Intelligence
  (IJCAI)}, 2021.

\bibitem{zou2019confidence}
Yang Zou, Zhiding Yu, Xiaofeng Liu, BVK Kumar, and Jinsong Wang.
\newblock Confidence regularized self-training.
\newblock In {\em ICCV}, 2019.

\bibitem{self:training:adaptation}
Yang Zou, Zhiding Yu, B.V.K. Vijaya~Kumar, and Jinsong Wang.
\newblock Unsupervised domain adaptation for semantic segmentation via
  class-balanced self-training.
\newblock In {\em ECCV}, 2018.

\end{thebibliography}
}

\begin{appendices}

\beginappendixa
\section{Mathematical Derivations}
\label{sec:extra_derivations}

\paragraph{Derivation of Log-Likelihood Loss}

We model the likelihood with an uncorrelated, bivariate Gaussian with mean $\hat{\mathbf{F}} = [\hat{F}^u, \hat{F}^v]^\top$ and variance $\hat{\Sigma} = \hat{\Sigma}^u = \hat{\Sigma}^v$ for flow directions $u$ and $v$.

\begin{equation}
    \begin{aligned}
        \mathcal{L}^{prob}_{\mathbf{I'}\rightarrow \mathbf{I}} =& -\log p(\mathbf{W} | \mathbf{I}, \mathbf{I}') \\
        =& -\log \left( \frac{1}{\sqrt{2\pi \hat{\Sigma}^u_{\mathbf{I'}\rightarrow \mathbf{I}}}} e^{-\frac{1}{2\hat{\Sigma}^u_{\mathbf{I'}\rightarrow \mathbf{I}}} \left(\hat{F}^u_{\mathbf{I'}\rightarrow \mathbf{I}} - W^u \right)^2} \cdot \right. \\
        &\left. \frac{1}{\sqrt{2\pi \hat{\Sigma}^v_{\mathbf{I'}\rightarrow \mathbf{I}}}} e^{-\frac{1}{2\hat{\Sigma}^v_{\mathbf{I'}\rightarrow \mathbf{I}}} (\hat{F}^v_{\mathbf{I'}\rightarrow \mathbf{I}} - W^v)^2} \right) \\
        =& -\log \left( \frac{1}{2\pi \hat{\Sigma}_{\mathbf{I'}\rightarrow \mathbf{I}}} e^{-\frac{1}{2\hat{\Sigma}_{\mathbf{I'}\rightarrow \mathbf{I}}} \left\|\hat{\mathbf{F}}_{\mathbf{I'}\rightarrow \mathbf{I}} - \mathbf{W} \right\|^2} \right) \\
        \propto& \frac{1}{2\hat{\Sigma}_{\mathbf{I'}\rightarrow \mathbf{I}}} \left\|\hat{\mathbf{F}}_{\mathbf{I'}\rightarrow \mathbf{I}} - \mathbf{W} \right\|^2 + \log \hat{\Sigma}_{\mathbf{I'}\rightarrow \mathbf{I}} \\
        =& \frac{1}{2\hat{\Sigma}_{\mathbf{I'}\rightarrow \mathbf{I}}} \mathcal{L}_{\mathbf{I'}\rightarrow \mathbf{I}} + \log \hat{\Sigma}_{\mathbf{I'}\rightarrow \mathbf{I}}
    \end{aligned}
\end{equation}

\paragraph{Derivation of Confidence Map}

We integrate the bivariate Gaussian density function over a circle with radius $r$ (subscripts are omitted).

\begin{equation}
\begin{aligned}
    P_{\mathcal{R}} &= p(\| \mathbf{F} - \hat{\mathbf{F}} \| \leq r) \\ 
    &= \int_0^{2\pi} \int_0^r \frac{1}{2\pi\hat{\Sigma}} e^{-\frac{1}{2\hat{\Sigma}}\rho^2} \rho d\rho d\phi \\
    &= 1 - \exp{\frac{-r^2}{2 \hat{\Sigma}}}
\end{aligned}
\end{equation}

\beginappendixb
\section{Training Details}
\label{sec:training_details}

In this section, we describe training settings and implementation details.
Both alignment and segmentation network were trained using Automatic Mixed Precision on a single consumer RTX 2080 Ti GPU.

\subsection{Alignment Network}
\label{subsec:alignment_training_details}

UAWarpC training almost exactly follows the setup of~\cite{truong2021warp}.
The training consists of two stages:
In the first stage, the network is trained without the visibility mask, as the visibility mask estimate is still inaccurate.
In the second stage, the visibility mask is activated and more data augmentation is used.

\paragraph{Data Handling}

The alignment network is trained using MegaDepth~\cite{li2018megadepth}, consisting of 196 scenes reconstructed from 1,070,468 internet photos with COLMAP~\cite{schonberger2016structure}. 
150 scenes are used for training, encompassing around 58,000 sampled image pairs.
1800 image pairs sampled from 25 different scenes are used for validation.
No ground-truth correspondences from SfM reconstructions are used to train UAWarpC.

During training, the image pairs $\mathbf{I}, \mathbf{J}$ are resized to 750\texttimes 750 pixels, and a dense flow $\mathbf{W}$ is sampled to create $\mathbf{I}'$.
Finally, all three images $\mathbf{I}, \mathbf{J}, \mathbf{I}'$ are center-cropped to resolution 520\texttimes 520.
In the first training stage, $\mathbf{W}$ consists of sampled color jitter, Gaussian blur, homography, TPS, and affine-TPS transformations.
In the second stage, local elastic transformations are added, and the strength of the transformations is increased.
For the detailed augmentation parameters, we refer to~\cite{truong2021warp}. % or our code.

\paragraph{Architecture and Loss Function}

Again following~\cite{truong2021warp}, a modified GLU-Net~\cite{truong2020glu} is used as a base architecture for flow prediction.
GLU-Net is a four-level pyramidal network with a VGG-16~\cite{simonyan2014very} encoder.
The encoder is initialized with ImageNet weights and frozen.
GLU-Net requires an additional low-resolution input of 256\texttimes 256 to establish global correlations, followed by repeated levels of upscaling and local feature correlations.
As in~\cite{truong2021warp}, our flow decoder uses residual connections for efficiency.
In addition, we replace all transposed convolutions with bilinear upsampling, and normalize all encoder feature maps, to increase the convergence rate.

The uncertainty estimate is produced using the uncertainty decoder proposed in~\cite{truong2021pdc}.
However, instead of predicting the parameters of several mixture components, we simply output a single value per pixel\textemdash the log-variance.

As in~\cite{truong2021warp}, the loss is applied at all four levels of the pyramidal GLU-Net.
We simply add the four components.
The employed loss functions are explained in Sec.~\ref{sec:method:alignment} of the main paper.
To obtain the visibility mask for the second training stage, we use the Cauchy-Schwarz inequality, analogously to~\cite{truong2021warp}.
\begin{equation}
\begin{split}
    V = \mathbf{1} \left[ \left\| \hat{\mathbf{F}}_{\mathbf{I}'\rightarrow \mathbf{J}} + \Phi_{\hat{\mathbf{F}}_{\mathbf{I}'\rightarrow \mathbf{J}}}(\hat{\mathbf{F}}_{\mathbf{J}\rightarrow \mathbf{I}}) - \mathbf{W} \right\|^2 < \right. \\ 
    \left. \alpha_2 + \alpha_1 \left( \left\| \hat{\mathbf{F}}_{\mathbf{I}'\rightarrow \mathbf{J}} \right\|^2 + \left\| \Phi_{\hat{\mathbf{F}}_{\mathbf{I}'\rightarrow \mathbf{J}}}(\hat{\mathbf{F}}_{\mathbf{J}\rightarrow \mathbf{I}}) \right\|^2 + \left\| \vphantom{\hat{\mathbf{F}}_{\mathbf{I}'\rightarrow \mathbf{J}}} \mathbf{W} \right\|^2 \right)  \right]
\end{split}
\end{equation}
$\mathbf{1}$ denotes the element-wise indicator function.
We use $\alpha_1 = 0.03$ and $\alpha_2 = 0.05$.

\paragraph{Optimization Schedule}

For the first training stage, the alignment network is trained with a batch size of 6 for 400k iterations.
We use the Adam optimizer~\cite{kingma2014adam} with weight decay $4\cdot10^{-4}$.
The initial learning rate is $10^{-4}$, and is halved after 250k and 325k iterations.
For the second training stage, we use 225k training steps with initial learning rate $5\cdot10^{-5}$, halved after 100k, 150k, and 200k iterations.

\subsection{Segmentation Network}

For training the domain adaptive segmentation network, we follow the employed base UDA method, respectively.
We summarize here the settings used with DAFormer~\cite{hoyer2021daformer}.
For more details, and the DACS~\cite{tranheden2021dacs} settings, we refer to the original papers or the authors' codes\footnote{\url{https://github.com/lhoyer/DAFormer}, \\  \url{https://github.com/vikolss/DACS}}.

\paragraph{Data Handling}

Input images are resized to half resolution for Cityscapes~\cite{cordts2016cityscapes}, ACDC~\cite{sakaridis2021acdc}, and Dark Zurich~\cite{sakaridis2020map}.
For RobotCar Correspondence~\cite{maddern20171,larsson2019cross} and CMU Correspondence~\cite{badino2011visual,larsson2019cross}, we resize to 720\texttimes 720 and 540\texttimes 720, respectively.
Data augmentation consists of random cropping to 512\texttimes 512 and random horizontal flipping.
For the coarsely labeled extra target images in the semi-supervised domain adaptation for RobotCar and CMU, we additionally apply random rotation with maximum 10° and color jittering.

\paragraph{Optimization Schedule}

We use the AdamW~\cite{loshchilov2017decoupled} optimizer with a weight decay of 0.01.
The learning rate follows a linear warmup for 1500 steps, followed by linear decay.
The peak learning rate is $6\cdot 10^{-4}$.
On ACDC and Dark Zurich, we train for 40k iterations; on RobotCar and CMU, we train for 20k iterations.
A batch size of 2 is used throughout.

To mitigate the risk of overfitting, we use the coarsely labeled extra target images in semi-supervised domain adaptation on RobotCar and CMU only in every second training iteration.

\beginappendixc
\section{Small \vs Large Static Classes}
\label{sec:large_static_extra}

\begin{figure}
    \begin{minipage}[c]{0.49\linewidth}
    \adjustbox{max width=\linewidth}{
        \input{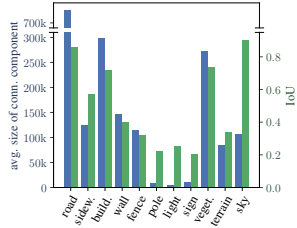}
    }
    \end{minipage}\hfill
    \begin{minipage}[c]{0.49\linewidth}
    \caption{Correlation between the average size of connected components (on Cityscapes~\cite{cordts2016cityscapes}) and mIoU score of warped reference image predictions for static classes. Larger classes benefit heavily from indiscriminate warping.}
    \label{fig:conn_comp}
    \end{minipage}
\end{figure}

\begin{figure}
    \begin{minipage}[c]{0.49\linewidth}
        \adjustbox{max width=\linewidth}{
            \input{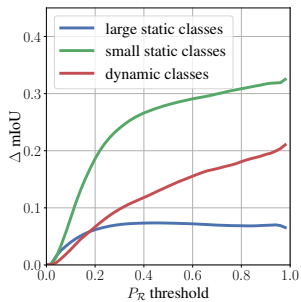}
        }
    \end{minipage}\hfill
    \begin{minipage}[c]{0.49\linewidth}
    \caption{Performance increase for different class categories as a function of the warp confidence ($P_\mathcal{R}$) threshold. Dynamic classes and small static classes (see Sec.~\ref{sec:method:refinement}) are more sensitive to the warp confidence, while large static classes do not improve considerably.}
    \label{fig:conf_thresh}
    \end{minipage}
\end{figure}

\beginappendixd
\begin{table*}
\caption{State-of-the-art comparison on Dark Zurich-test for Cityscapes\textrightarrow Dark Zurich domain adaptation. Methods above the double line all use a DeepLabv2~\cite{chen2017deeplab} model. ``Ref.'': For each adverse input image a reference image at similar geo-location is used.}
\smallskip%
\centering
\resizebox{\linewidth}{!}{%
\ra{1.3}%
\begin{tabular}{@{}lcccccccccccccccccccccccr@{}}\toprule
\multirow{2}{*}{Method} && \multirow{2}{*}{Ref.} && \multicolumn{21}{c}{IoU\,$\uparrow$} \\
\cmidrule{5-25} &&&& \rotatebox[origin=c]{90}{road} & \rotatebox[origin=c]{90}{sidew.} & \rotatebox[origin=c]{90}{build.} & \rotatebox[origin=c]{90}{wall} & \rotatebox[origin=c]{90}{fence} & \rotatebox[origin=c]{90}{pole} & \rotatebox[origin=c]{90}{light} & \rotatebox[origin=c]{90}{sign} & \rotatebox[origin=c]{90}{veget.} & \rotatebox[origin=c]{90}{terrain} & \rotatebox[origin=c]{90}{sky} & \rotatebox[origin=c]{90}{person} & \rotatebox[origin=c]{90}{rider} & \rotatebox[origin=c]{90}{car} & \rotatebox[origin=c]{90}{truck} & \rotatebox[origin=c]{90}{bus} & \rotatebox[origin=c]{90}{train} & \rotatebox[origin=c]{90}{motorc.} & \rotatebox[origin=c]{90}{bicycle} && \multicolumn{1}{c}{\phantom{00}\rotatebox[origin=c]{90}{\textbf{mean}}} \\ \midrule
DeepLabv2~\cite{chen2017deeplab} &&&& 79.0 & 21.8 & 53.0 & 13.3 & 11.2 & 22.5 & 20.2 & 22.1 & 43.5 & 10.4 & 18.0 & 37.4 & 33.8 & 64.1 & \phantom{0}6.4 & \phantom{0}0.0 & 52.3 & 30.4 & \phantom{0}7.4 && 28.8 \\
ADVENT~\cite{vu2019advent} &&&& 85.8 & 37.9 & 55.5 & 27.7 & 14.5 & 23.1 & 14.0 & 21.1 & 32.1 & \phantom{0}8.7 & \phantom{0}2.0 & 39.9 & 16.6 & 64.0 & 13.8 & \phantom{0}0.0 & 58.8 & 28.5 & 20.7 && 29.7 \\
AdaptSegNet~\cite{tsai2018learning} &&&& 86.1 & 44.2 & 55.1 & 22.2 & 4.8 & 21.1 & \phantom{0}5.6 & 16.7 & 37.2 & \phantom{0}8.4 & \phantom{0}1.2 & 35.9 & 26.7 & 68.2 & 45.1 & \phantom{0}0.0 & 50.1 & 33.9 & 15.6 && 30.4 \\
BDL~\cite{li2019bidirectional} &&&& 85.3 & 41.1 & 61.9 & 32.7 & 17.4 & 20.6 & 11.4 & 21.3 & 29.4 & \phantom{0}8.9 & \phantom{0}1.1 & 37.4 & 22.1 & 63.2 & 28.2 & \phantom{0}0.0 & 47.7 & 39.4 & 15.7 && 30.8 \\
DANNet (DeepLabv2)~\cite{wu2021dannet} && \checkmark && 88.6 & 53.4 & 69.8 & 34.0 & 20.0 & 25.0 & 31.5 & 35.9 & 69.5 & 32.2 & 82.3 & 44.2 & 43.7 & 54.1 & 22.0 & \phantom{0}0.1 & 40.9 & 36.0 & 24.1 && 42.5 \\
DANIA (DeepLabv2)~\cite{wu2021one} && \checkmark && 89.4 & 60.6 & 72.3 & 34.5 & 23.7 & 37.3 & 32.8 & 40.0 & 72.1 & 33.0 & 84.1 & 44.7 & 48.9 & 59.0 & \phantom{0}9.8 & \phantom{0}0.1 & 40.1 & 38.4 & 30.5 && 44.8 \\
% SePiCo (DeepLabv2)~\cite{xie2022sepico} && \checkmark && 91.2 & 61.3 & 67.0 & 28.5 & 15.5 & 44.7 & 44.3 & 41.3 & 65.4 & 22.5 & 80.4 & 41.3 & 52.4 & 71.2 & 39.3 & \phantom{0}0.0 & 39.6 & 27.5 & 28.8 && 45.4 \\
\midrule
DACS~\cite{tranheden2021dacs} &&&& 83.1 & 49.1 & 67.4 & 33.2 & 16.6 & 42.9 & 20.7 & 35.6 & 31.7 & \phantom{0}5.1 & \phantom{0}6.5 & 41.7 & 18.2 & 68.8 & \textbf{76.4} & \phantom{0}0.0 & 61.6 & 27.7 & 10.7 && 36.7 \\
% DACS w/ Test-Refign && \checkmark && 88.8 & 57.8 & 76.1 & 38.2 & 22.5 & 43.7 & 23.9 & 37.5 & 49.6 & 11.1 & 62.1 & 41.4 & 17.6 & 69.4 & \textbf{80.8} & \phantom{0}0.0 & 62.1 & 28.1 & 14.0 && 43.4 \\
Refign-DACS && \checkmark && 89.9 & 59.7 & 69.5 & 28.5 & 11.6 & 39.0 & 17.1 & 35.0 & 35.7 & 18.8 & 30.4 & 38.8 & 43.1 & 72.3 & 73.7 & \phantom{0}0.0 & 61.6 & 33.9 & 24.7 && 41.2 \\
\midrule\midrule
DMAda (RefineNet)~\cite{dai2018dark} && \checkmark && 75.5 & 29.1 & 48.6 & 21.3 & 14.3 & 34.3 & 36.8 & 29.9 & 49.4 & 13.8 & \phantom{0}0.4 & 43.3 & 50.2 & 69.4 & 18.4 & \phantom{0}0.0 & 27.6 & 34.9 & 11.9 && 32.1 \\
GCMA (RefineNet)~\cite{sakaridis2019guided} && \checkmark && 81.7 & 46.9 & 58.8 & 22.0 & 20.0 & 41.2 & 40.5 & 41.6 & 64.8 & 31.0 & 32.1 & 53.5 & 47.5 & 75.5 & 39.2 & \phantom{0}0.0 & 49.6 & 30.7 & 21.0 && 42.0 \\
MGCDA (RefineNet)~\cite{sakaridis2020map} && \checkmark && 80.3 & 49.3 & 66.2 & \phantom{0}7.8 & 11.0 & 41.4 & 38.9 & 39.0 & 64.1 & 18.0 & 55.8 & 52.1 & \textbf{53.5} & 74.7 & 66.0 & \phantom{0}0.0 & 37.5 & 29.1 & 22.7 && 42.5 \\
CDAda (RefineNet)~\cite{xu2021cdada} && \checkmark && 90.5 & 60.6 & 67.9 & 37.0 & 19.3 & 42.9 & 36.4 & 35.3 & 66.9 & 24.4 & 79.8 & 45.4 & 42.9 & 70.8 & 51.7 & \phantom{0}0.0 & 29.7 & 27.7 & 26.2 && 45.0 \\
DANNet (PSPNet)~\cite{wu2021dannet} && \checkmark && 90.4 & 60.1 & 71.0 & 33.6 & 22.9 & 30.6 & 34.3 & 33.7 & 70.5 & 31.8 & 80.2 & 45.7 & 41.6 & 67.4 & 16.8 & \phantom{0}0.0 & 73.0 & 31.6 & 22.9 && 45.2 \\
% Bi-Mix (RefineNet)~\cite{yang2021bi} && \checkmark && 89.2 & 59.4 & 75.8 & 41.7 & 19.2 & 39.0 & 31.9 & 31.5 & 70.9 & 30.1 & 81.9 & 44.9 & 41.8 & 66.3 & 34.2 & \phantom{0}1.0 & 61.1 & 47.4 & 14.6 && 46.5 \\
CCDistill (RefineNet)~\cite{gao2022cross} && \checkmark && 89.6 & 58.1 & 70.6 & 36.6 & 22.5 & 33.0 & 27.0 & 30.5 & 68.3 & 33.0 & 80.9 & 42.3 & 40.1 & 69.4 & 58.1 & \phantom{0}0.1 & 72.6 & \textbf{47.7} & 21.3 && 47.5 \\
DANIA (PSPNet)~\cite{wu2021one} && \checkmark && 91.5 & 62.7 & 73.9 & \textbf{39.9} & 25.7 & 36.5 & 35.7 & 36.2 & 71.4 & \textbf{35.3} & 82.2 & 48.0 & 44.9 & 73.7 & 11.3 & \phantom{0}0.1 & 64.3 & 36.7 & 22.7 && 47.0 \\
% SePiCo (DAFormer)~\cite{xie2022sepico} && \checkmark && \textbf{93.2} & \textbf{68.1} & 73.7 & 32.8 & 16.3 & 54.6 & \textbf{49.5} & 48.1 & 74.2 & 31.0 & 86.3 & 57.9 & 50.9 & \textbf{82.4} & 52.2 & \phantom{0}1.3 & 83.8 & 43.9 & 29.8 && 54.2 \\
\midrule
DAFormer~\cite{hoyer2021daformer} &&&& \textbf{93.5} & \textbf{65.5} & 73.3 & 39.4 & 19.2 & 53.3 & 44.1 & 44.0 & 59.5 & 34.5 & 66.6 & 53.4 & 52.7 & \textbf{82.1} & 52.7 & \textbf{\phantom{0}9.5} & 89.3 & \textbf{50.5} & \textbf{38.5} && 53.8 \\
% DAFormer w/ Test-Refign && \checkmark && 92.9 & 67.0 & 75.1 & \textbf{42.8} & 24.1 & 50.6 & 35.3 & 42.1 & 65.0 & 33.6 & 73.3 & 54.6 & 53.0 & 79.0 & 75.1 & \phantom{0}0.0 & 90.2 & \textbf{48.3} & \textbf{40.8} && 54.9 \\
Refign-DAFormer && \checkmark && 91.8 & 65.0 & \textbf{80.9} & 37.9 & \textbf{25.8} & \textbf{56.2} & \textbf{45.2} & \textbf{51.0} & \textbf{78.7} & 31.0 & \textbf{88.9} & \textbf{58.8} & 52.9 & 77.8 & 51.8 & \phantom{0}6.1 & \textbf{90.8} & 40.2 & 37.1 && \textbf{56.2} \\
\bottomrule
\end{tabular}%
}
\label{tab:darkzurich_sota}
\end{table*}

To motivate the distinction between small and large static classes (as defined in Sec.~\ref{sec:method:refinement}), we generate ACDC~\cite{sakaridis2021acdc} reference image predictions using a SegFormer~\cite{xie2021segformer} trained on Cityscapes~\cite{cordts2016cityscapes}, and warp them onto the corresponding adverse-image viewpoint.
As shown in Fig.~\ref{fig:conn_comp}, we observe a correlation between the resulting IoU and the average size of the connected class component for static classes (pearson correlation coeff.\ of 0.70).
The classes \emph{pole}, \emph{traffic light}, and \emph{traffic sign} are drastically smaller than the rest, and consequentially have lower accuracy.
On the other hand, such indiscriminate warping (\ie, without $P_{\mathcal{R}}$) is surprisingly accurate for the large static classes.

Furthermore, we analyze the mIoU improvement when only considering pixels above a certain $P_{\mathcal{R}}$ threshold for the above mentioned warped SegFormer predictions, see Fig.~\ref{fig:conf_thresh}.
While the performance increases monotonically for both dynamic and small static classes, it remains mostly flat for large static classes.
This suggests that large static classes are largely insensitive to the warping confidence, while both dynamic and small static classes benefit greatly from confidence guidance.

\section{Additional Experimental Results}

\begin{table}
\caption{State-of-the-art comparison of models which do not follow the common image input resizing protocol. Refign-HRDA currently ranks first on public leaderboards.}
\smallskip%
\centering
\resizebox{\linewidth}{!}{%
\ra{1.3}%
\begin{tabular}{@{}lccccr@{}}\toprule
\multirow{2}{*}{Method} & \multicolumn{1}{c}{Cityscapes\textrightarrow ACDC} && \multicolumn{3}{c}{Cityscapes\textrightarrow Dark Zurich} \\
\cmidrule{2-2} \cmidrule{4-6} & ACDC~\cite{sakaridis2021acdc} && Dark Zurich-test~\cite{sakaridis2020map} & ND~\cite{dai2018dark} & Bn~\cite{yu2020bdd100k,sakaridis2020map} \\
\midrule
SePiCo (DAFormer)~\cite{xie2022sepico} & - && 54.2 & 57.1 & 36.9 \\
HRDA~\cite{hoyer2022hrda} & 68.0 && 55.9 & 55.6 & 39.1 \\
\midrule
Refign-HRDA & \textbf{72.1} && \textbf{63.9} & \textbf{57.8} & \textbf{40.6} \\
\bottomrule
\end{tabular}%
}
\label{tab:larger_input}
\end{table}

Due to space restrictions, we present the full class-wise performances of state-of-the-art UDA methods on Dark Zurich-test here in Table~\ref{tab:darkzurich_sota}.
The models reported in Tables~\ref{tab:acdc_sota}, \ref{tab:night_generalization}, and \ref{tab:darkzurich_sota} all use the same image input size at test-time for fairness of comparison.
Table~\ref{tab:larger_input} presents models which do not follow that protocol.
Using Cityscapes-pretrained weights for initialization, Refign added on top of HRDA~\cite{hoyer2022hrda} achieves 72.1\,mIoU and 63.9\,mIoU on ACDC and Dark Zurich-test, respectively, ranking first on the public leaderboards of these benchmarks at the time of publication.

In Table~\ref{tab:conditionwise}, we report the performance of Cityscapes\textrightarrow ACDC Refign-DAFormer on the four different conditions of the ACDC validation set. Refign improves markedly over the baseline for all conditions.

\begin{table}
\caption{Performance of Cityscapes\textrightarrow ACDC models for different conditions on the validation set.}%
\smallskip%
\centering%
\resizebox{\linewidth}{!}{%
    \ra{1.3}%
\setlength{\tabcolsep}{22pt}%
\begin{tabular}{@{}lccccr@{}}\toprule
\multirow{2}{*}{Method} && \multicolumn{4}{c}{mIoU\,$\uparrow$} \\
\cmidrule{3-6} && night & snow & rain & fog \\
\midrule
DAFormer~\cite{hoyer2021daformer} && 34.8 & 56.3 & 58.5 & 67.9 \\
Refign-DAFormer && \textbf{48.1} & \textbf{65.0} & \textbf{65.2} & \textbf{73.4} \\
\bottomrule
\end{tabular}%
    }%
\label{tab:conditionwise}
\end{table}

We also compare the Cityscapes\textrightarrow ACDC Refign-DAFormer model with state-of-the-art foggy scene understanding methods in Table~\ref{tab:fog_generalization}.
All methods are trained with Cityscapes as source domain, however the foggy scene understanding methods utilize both synthetic foggy data and a larger pool of real foggy data as targets.
Surprisingly, our model achieves state-of-the-art performance despite this handicap.

\begin{table}
\caption{Performance comparison with specialized foggy scene understanding methods on the Foggy Zurich~\cite{dai2020curriculum} and Foggy Driving~\cite{sakaridis2018semantic} test sets.}
\smallskip%
\centering
\resizebox{\columnwidth}{!}{%
\ra{1.3}%
\begin{tabular}{@{}lccccccr@{}}\toprule
\multirow{2}{*}{Method} && \multicolumn{3}{c}{Target Domain Training Data} && \multicolumn{2}{c}{mIoU\,$\uparrow$} \\
\cmidrule{3-5} \cmidrule{7-8} && Foggy CS-DBF~\cite{dai2020curriculum} & Foggy Zurich\cite{dai2020curriculum} & ACDC~\cite{sakaridis2021acdc} && Foggy Zurich~\cite{dai2020curriculum} & FoggyDriving~\cite{sakaridis2018semantic}\\
\midrule
CMAda3+~\cite{dai2020curriculum} && \checkmark & \checkmark &&& 46.8 & 49.8 \\
FIFO~\cite{lee2022fifo} && \checkmark & \checkmark &&& 48.4 & 50.7 \\
CuDA-Net+~\cite{ma2021both} && \checkmark & \checkmark &&& 49.1 & 53.5 \\
TDo-Dif~\cite{liao2022unsupervised} && \checkmark & \checkmark &&& \textbf{51.9} & 50.7 \\
\midrule
Refign-DAFormer && && \checkmark && 51.4 & \textbf{53.9} \\
\bottomrule
\end{tabular}%
}
\label{tab:fog_generalization}
\end{table}

Finally, we conduct experiments substituting the SegFormer~\cite{xie2021segformer} based architecture of DAFormer~\cite{hoyer2021daformer} with DeepLabv2~\cite{chen2017deeplab}.
On both ACDC and Dark Zurich validation sets, this version of Refign improves substantially over the baseline, as reported in Table~\ref{tab:daformer_deeplabv2}.

\begin{table}
\caption{Performance of Refign \vs DAFormer baseline with a DeepLabv2 model on the ACDC and Dark Zurich validation sets.}
\smallskip%
\centering
\resizebox{\columnwidth}{!}{%
\ra{1.3}%
\setlength{\tabcolsep}{20pt}%
\begin{tabular}{@{}lccr@{}}\toprule
\multirow{2}{*}{Method} && \multicolumn{2}{c}{mIoU\,$\uparrow$} \\
\cmidrule{3-4} && ACDC~\cite{sakaridis2021acdc} & Dark Zurich~\cite{sakaridis2020map}\\
\midrule
DAFormer (DeepLabv2)~\cite{hoyer2021daformer} && 46.4 & 24.8 \\
Refign-DAFormer (DeepLabv2) && \textbf{55.6} & \textbf{38.7} \\
\bottomrule
\end{tabular}%
}
\label{tab:daformer_deeplabv2}
\end{table}

\beginappendixe
\section{Refign at Test-Time}

Although designed to refine pseudo-labels during online self-training, Refign can also be applied at test-time to arbitrary, trained models, if a reference image is available.
We report ACDC and Dark Zurich validation set scores in Table~\ref{tab:test_refign}.
The performance gain is more moderate than if Refign is applied at training-time.
This is unsurprising, given that we only conduct a single refinement iteration in that case.

\begin{table}
\caption{Applying Refign only for one refinement iteration at test-time to DAFormer on the ACDC and Dark Zurich validation sets.}
\smallskip%
\centering
\resizebox{\columnwidth}{!}{%
\ra{1.3}%
\setlength{\tabcolsep}{20pt}%
\begin{tabular}{@{}lccr@{}}\toprule
\multirow{2}{*}{Method} && \multicolumn{2}{c}{mIoU\,$\uparrow$} \\
\cmidrule{3-4} && ACDC~\cite{sakaridis2021acdc} & Dark Zurich~\cite{sakaridis2020map}\\
\midrule
DAFormer~\cite{hoyer2021daformer} && 55.6 & 34.1 \\
DAFormer + Test-Time Refign && \textbf{56.8} & \textbf{38.0} \\
\bottomrule
\end{tabular}%
}
\label{tab:test_refign}
\end{table}

\beginappendixf
\section{Qualitative Results}

\begin{figure*}
    \adjustbox{max width=\textwidth}{
        \input{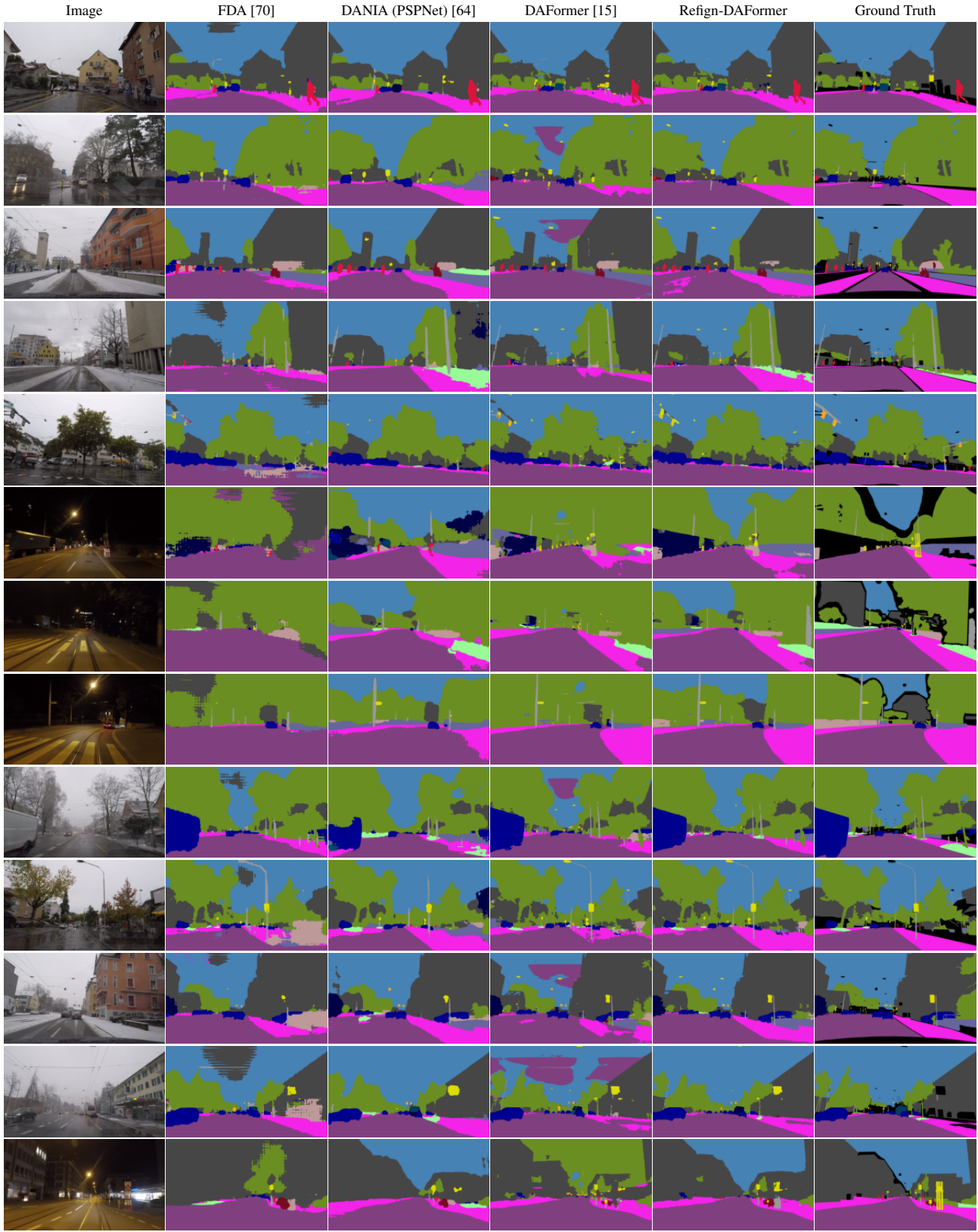}
    }
    \caption{Prediction samples of the ACDC validation set.}
    \label{fig:preds_supp}
\end{figure*}

\begin{figure*}
    \adjustbox{max width=\textwidth}{
        \input{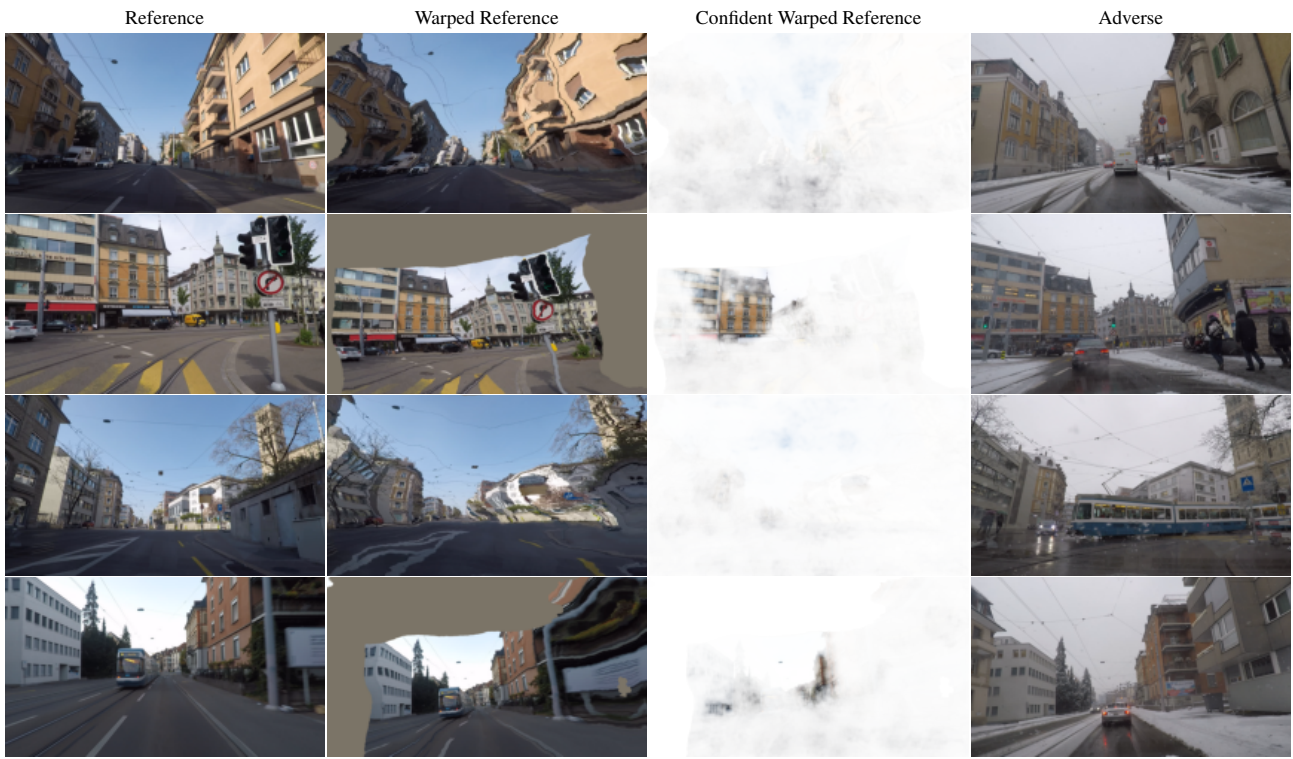}
    }
    \caption{Warp failure examples on ACDC.}
    \label{fig:warped_failures}
\end{figure*}

We show more qualitative results in this section.
Fig.~\ref{fig:preds_supp} shows more qualitative segmentation results for randomly selected ACDC validation samples.
Fig.~\ref{fig:warped_images_supp} shows the warps and corresponding confidence maps for randomly selected ACDC samples.
Finally, in Fig.~\ref{fig:warped_failures}, we show some warp failures.
Importantly, the confidence map correctly blends out the inaccurate warps.

\begin{figure*}
    \adjustbox{max width=\textwidth}{
        \input{figures/warped_images_full.pgf}
    }
    \caption{Example visualizations of warped reference images and the corresponding confidence maps from ACDC.}
    \label{fig:warped_images_supp}
\end{figure*}

\end{appendices}

\end{document}